\documentclass[10pt,twocolumn,letterpaper]{article}

\usepackage[pagenumbers]{cvpr} 

\usepackage{algorithm}
\usepackage{algpseudocode}
\usepackage{comment}
\usepackage{multirow}
\usepackage{graphicx}
\usepackage{subcaption}
\usepackage[table,xcdraw]{xcolor}
\usepackage[accsupp]{axessibility} 

\definecolor{cvprblue}{rgb}{0.21,0.49,0.74}
\usepackage[pagebackref,breaklinks,colorlinks,allcolors=blue]{hyperref}

\def\paperID{***} 
\def\confName{CVPR}
\def\confYear{2026}

\title{Recover to Predict: Progressive Retrospective Learning for Variable-Length Trajectory Prediction}

\author{
Hao Zhou$^{1,2,*}$,~~~
Lu Qi$^{3,*}$,~~~
Jason Li$^{4}$,~~~
Jie Zhang$^1$,~~~ \\
Yi Liu$^{5}$,~~~
Xu Yang$^6$, ~~~
Mingyu Fan$^{5,\dagger}$, ~~~
Fei Luo$^{1,\dagger}$%
\begingroup

\thanks{~Equal contribution. ${}^\dagger$ Co-corresponding authors.}
\endgroup
\\[0.2cm]
$^1$Great Bay University~~
$^2$Tsinghua SIGS~~
$^3$Wuhan University~~
$^4$NTU~~
$^5$Donghua University~~
$^6$CASIA
}

\begin{document}
\maketitle
\begin{abstract}

Trajectory prediction is critical for autonomous driving, enabling safe and efficient planning in dense, dynamic traffic. 
Most existing methods optimize prediction accuracy under fixed-length observations.
However, real-world driving often yields variable-length, incomplete observations, posing a challenge to these methods.
%
A common strategy is to directly map features from incomplete observations to those from complete ones.
%
This one-shot mapping, however, struggles to learn accurate representations for short trajectories due to significant information gaps.
%
To address this issue, we propose a \textbf{P}rogressive \textbf{R}etrospective \textbf{F}ramework (PRF), which gradually aligns features from incomplete observations with those from complete ones via a cascade of retrospective units. 
Each unit consists of a Retrospective Distillation Module (RDM) and a Retrospective Prediction Module (RPM), where RDM distills features and RPM recovers previous timesteps using the distilled features. 
Moreover, we propose a Rolling-Start Training Strategy (RSTS) that enhances data efficiency during PRF training. 
%
PRF is plug-and-play with existing methods.
Extensive experiments on datasets Argoverse~2 and Argoverse~1 demonstrate the effectiveness of PRF.
Code is available at \url{https://github.com/zhouhao94/PRF}.
\end{abstract}    
\vspace{-4mm}
\section{Introduction}
\vspace{-1mm}
\label{sec:intro}

\begin{figure}[!t]
  \centering

  \begin{subfigure}[t]{0.38\linewidth}
    \centering
    \includegraphics[width=\linewidth]{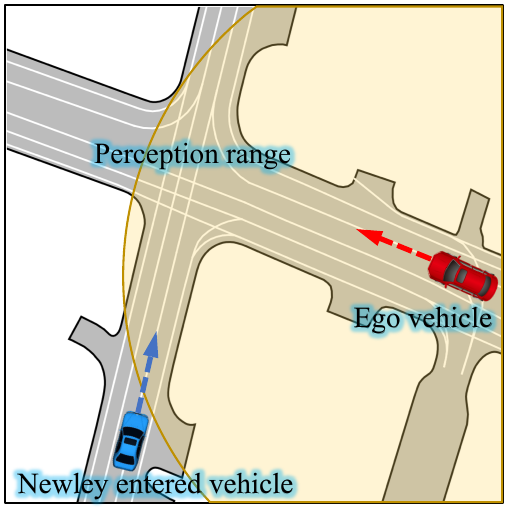}
    \caption{Newly entered vehicle}\label{fig:teaser-a}
  \end{subfigure}\hspace{2.5mm}
  \begin{subfigure}[t]{0.38\linewidth}
    \centering
    \includegraphics[width=\linewidth]{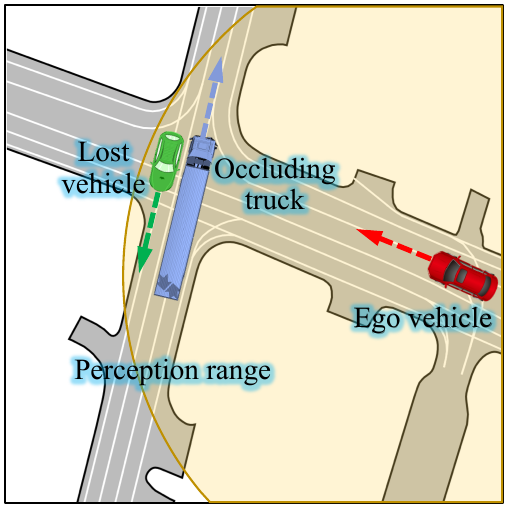}
    \caption{Tracking lost vehicle}\label{fig:teaser-b}
  \end{subfigure}

  \medskip
  \vspace{-3pt}

  \begin{subfigure}[t]{0.4\linewidth}
    \centering
    \includegraphics[width=\linewidth,trim={14pt 15pt 10pt 15pt},clip]{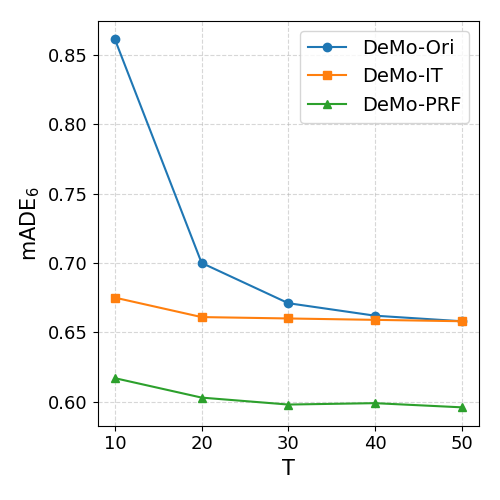}
    \vspace{-0.5mm}
    \caption{mADE$_6$ on Argoverse~2}\label{fig:teaser-c}
  \end{subfigure}\hspace{2pt}
  \begin{subfigure}[t]{0.4\linewidth}
    \centering
    \includegraphics[width=\linewidth,trim={14pt 15pt 10pt 15pt},clip]{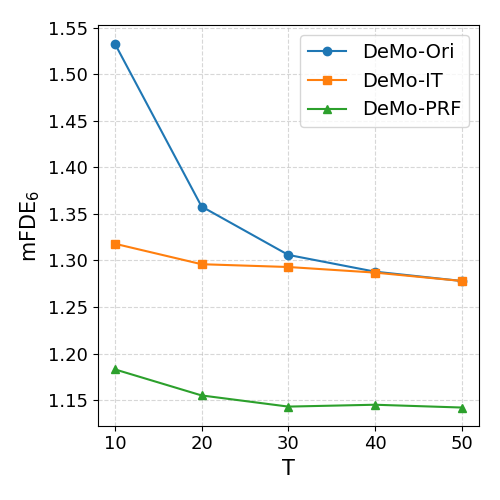}
    \vspace{-0.5mm}
    \caption{mFDE$_6$ on Argoverse~2}\label{fig:teaser-d}
  \end{subfigure}
  \vspace{-3mm}
  \caption{Fig.~\ref{fig:teaser-a} and Fig.~\ref{fig:teaser-b} display two common scenarios that yield variable-length, incomplete trajectories. Fig.~\ref{fig:teaser-c} and Fig.~\ref{fig:teaser-d} respectively present the mADE$_6$ and mFDE$_6$ results for the original DeMo~\cite{zhang2024decoupling}, DeMo with Isolated Training (DeMo-IT), and DeMo with PRF (DeMo-PRF) under varying observation lengths.}\label{fig:teaser}
  \vspace{-6mm}
\end{figure}

Trajectory prediction for dynamic agents in traffic scenarios is crucial for autonomous driving systems, enabling vehicles to anticipate the future motions of road users, plan safe and efficient maneuvers, and avoid collisions. 
The numerous learning-based methods~\cite{gao2020vectornet,liang2020learning,varadarajan2022multipath++,shi2022motion,zhou2022hivt,nayakanti2023wayformer,zhou2023query,zhou2024smartrefine,zhang2024decoupling,huang2025trajectory} have been proposed to improve prediction accuracy. 
Although these methods have made significant progress, they primarily focus on optimizing network architectures to improve prediction accuracy using \textit{idealized standard-length observations} as inputs. 

However, complete historical observations are often unavailable in real-world settings. 
For example, when a vehicle first enters the ego vehicle's perception range (Fig.~\ref{fig:teaser-a}) or is detected again after being lost due to occlusion or tracking errors (Fig.~\ref{fig:teaser-b}), the temporal context is insufficient to reconstruct a complete historical trajectory. Such incomplete, variable-length observations pose a challenge for existing methods. 
As shown in Fig.~\ref{fig:teaser-c} and Fig.~\ref{fig:teaser-d}, the performance of the state-of-the-art method DeMo~\cite{zhang2024decoupling} degrades significantly as the number of observed timesteps decreases. 
This degradation can propagate to downstream planning and control, increasing the risk of unsafe maneuvers and collisions in real-world driving.

A common, straightforward strategy for variable-length prediction is Isolate Training (IT).
It trains a separate model for each observation length and evaluates each model on inputs of the same length. 
Although IT yields modest gains on variable-length prediction, as illustrated in Fig.~\ref{fig:teaser-c} and Fig.~\ref{fig:teaser-d}, it incurs substantial computational and memory overhead because it requires training and maintaining multiple models across observation lengths. To improve both efficiency and performance, several learning-based methods~\cite{monti2022many,sun2022human,li2023bcdiff,xu2024adapting,qiu2025adapting} have been proposed. 
The key insight of these methods is to directly map features from variable-length observations to a canonical representation, typically aligned with either complete observations or a designated target length. 
This one-shot mapping strategy is relatively effective when observations are close to standard length, but it struggles to learn faithful representations for short trajectories due to pronounced information gaps.

In this work, we propose a new method, \textbf{P}rogressive \textbf{R}etrospective \textbf{F}ramework (PRF) for variable-length trajectory prediction. 
Instead of directly mapping features from incomplete to complete observations, PRF progressively aligns them via a cascade of retrospective units. 
%
This decomposition reduces the learning difficulty, as each unit only needs to bridge a small feature gap over a short temporal horizon.
Each unit consists of a Retrospective Distillation Module (RDM) and a Retrospective Prediction Module (RPM). RDM distills features of an incomplete trajectory to its previous history timesteps, while RPM reconstructs these missing timesteps from the distilled feature. 
PRF operates between the encoder and decoder, making it plug-and-play with existing approaches.
Fig.~\ref{fig:teaser-c} and Fig.~\ref{fig:teaser-d} show that PRF yields significant improvements across observation lengths using a single model trained once.

Since a shared encoder extracts features for variable-length observations, naïve distillation may lead to feature conflicts. 
Therefore, RDM adopts a \textit{residual-based} distillation strategy that models features at omitted timesteps as learnable residuals. 
RPM employs a decoupled query design that integrates anchor-free and anchor-based formulations, enabling coarse-to-fine historical retrospection. 
%
It provides implicit supervision for RDM's distillation.
%
Moreover, since each unit targets a specific observation length, incomplete observations can be used to train all units whose target lengths they cover. Accordingly, we propose a Rolling-Start Training Strategy (RSTS) to generate multiple samples from one sequence, improving data efficiency. 
%

The main contributions are as follows:
\begin{itemize}
    \item We design a Progressive Retrospective Framework~(PRF) for variable-length prediction. PRF progressively aligns features from variable-length observations with those from complete ones via a cascade of retrospective units.
    \item We propose a Retrospective Prediction Module (RPM) and a Retrospective Distillation Module (RDM) to form each unit jointly. RPM distills features, while RDM recovers the omitted history using the distilled features.
    \item We introduce a Rolling-Start Training Strategy (RSTS) to generate multiple training samples from a single sequence, enhancing data efficiency for training PRF.
    \item We perform extensive experiments on Argoverse~2 and Argoverse~1, demonstrating that PRF significantly improves variable-length prediction and achieves state-of-the-art results on standard benchmarks.
\end{itemize}
\vspace{-2mm}
\section{Related Work}
\vspace{-1mm}
\label{sec:related_work}

\textbf{Trajectory Prediction.} In autonomous driving, scene representation is crucial for accurate prediction. 
Traditional methods~\cite{phan2020covernet,gilles2021home,chai2020multipath} rasterize driving scenarios and use CNNs for context extraction. However, CNNs struggle to capture scenario details, motivating vectorized scene representations~\cite{zhao2021tnt,gu2021densetnt,zhou2022hivt,shi2024mtr++} first introduced by VectorNet~\cite{gao2020vectornet}. Based on vectorization, attention~\cite{li2020end,liu2021multimodal,mercat2020multi} and graph convolutions~\cite{gilles2022gohome,jia2023hdgt,rowe2023fjmp,tang2024hpnet} have been widely explored to model agent-scene interactions. Conditioned on the scene encoding, numerous methods have been proposed for multimodal trajectory prediction. Early works adopt goal-conditioned strategies~\cite{zhao2021tnt,gu2021densetnt,liu2024reasoning} or probability distribution heatmaps~\cite{gilles2021home,gilles2022gohome}. Recently, with the rise of the Transformer~\cite{vaswani2017attention}, Transformer-based models~\cite{liu2021multimodal,ngiamscene,shi2022motion,nayakanti2023wayformer,huang2023gameformer}, such as QCNet~\cite{zhou2023query} and DeMo~\cite{zhang2024decoupling}, have become the dominant paradigm. Moreover, techniques including pre-training~\cite{chen2023traj,cheng2023forecast,lan2024sept}, post-refinement~\cite{choi2023r,zhou2024smartrefine}, GPT~\cite{philion2023trajeglish,seff2023motionlm}, Diffusion~\cite{jiang2023motiondiffuser}, and Mamba~\cite{huang2025trajectory} have further advanced performance. However, these methods show limitations when using variable-length observations as inputs. 

\begin{figure*}[!t]
    \centering
    \includegraphics[width=0.85\textwidth]{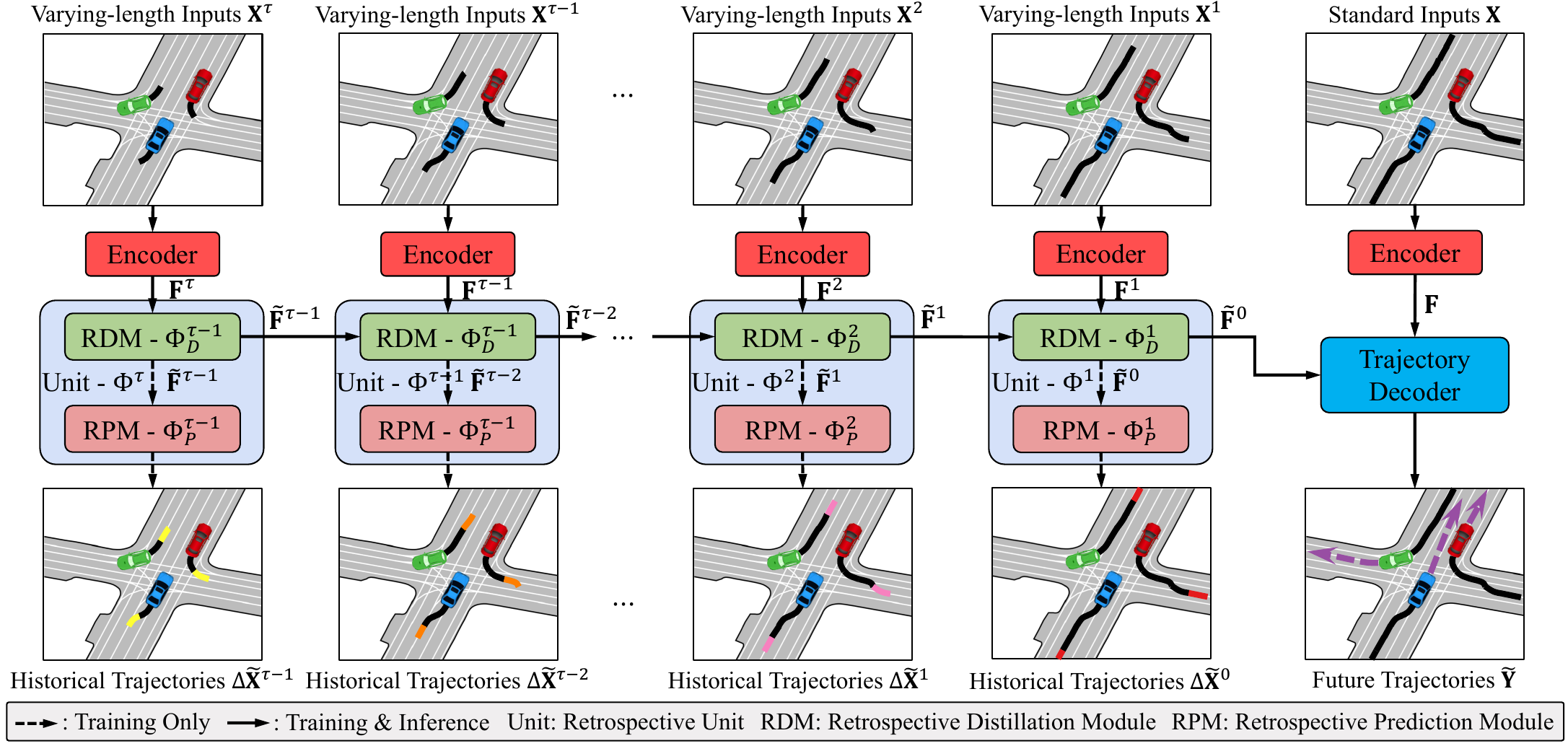}
    \vspace{-2mm}
    \caption{Overview of the PRF. A cascade of retrospective units progressively distills features of varying-length inputs, aligning them with those from complete ones to improve prediction performance. Each unit includes a Retrospective Distillation Module (RDM) that distills features to longer observations and a Retrospective Prediction Module (RPM) that recovers omitted history from the distilled features.}
    \label{fig:arch}
    \vspace{-5mm}
\end{figure*}

\noindent \textbf{Variable-Length Trajectory Prediction.} Incomplete and variable-length trajectories are common in real-world applications and have attracted increasing attention. DTO~\cite{monti2022many} distills knowledge from a teacher trained on complete trajectories to a student that predicts from short inputs. MOE~\cite{sun2022human} introduces a feature extractor for momentary observations and a pre-training scheme that recovers observations and context. BCDiff~\cite{li2023bcdiff} develops two coupled diffusions to infer historical and future trajectories from limited observations. FLN~\cite{xu2024adapting} designs calibration and adaptation modules to learn temporally invariant representations. LaKD~\cite{li2024lakd} proposes length-agnostic knowledge distillation to transfer knowledge across different observation lengths. CLLS~\cite{qiu2025adapting} employs contrastive learning to extract length-invariant features. 
Despite notable advances, these methods directly map variable-length observations to a canonical representation. This works for near-standard inputs but often fails on short trajectories due to large information gaps.
Our PRF progressively aligns them by a cascade of units, thereby reducing learning difficulty.

\vspace{-2mm}
\section{Method}
\vspace{-1mm}
\label{sec:method}


\subsection{Problem Formulation} 
\label{sec.3-1}

In a driving scenario, the vectorized map is denoted by $\mathbf{M} \in \mathbb{R}^{P \times S \times C_m} $, where $P$, $S$, and $C_m$ are the number of map polylines, divided segments, and feature channels, respectively. 
The observed trajectories of agents are represented by $\mathbf{X} \in \mathbb{R}^{N \times T_o \times C_a}$, where $N$, $T_o$, and $C_a$ are the number of agents, observed timesteps, and motion states (\textit{e.g.}, position, heading angle, velocity). 
The future trajectories of the target agents are represented by $ \mathbf{Y} \in \mathbb{R}^{N_a \times T_f \times 2}$, where $N_a$ is the number of selected agents and $T_f$ is the prediction horizon. 
The standard trajectory prediction task is to learn a generative method $p_\theta(\mathbf{Y|\mathbf{X},\mathbf{M}})$, that predicts future trajectories $\mathbf{Y}$ based on the observed trajectories $\mathbf{X}$ and the vectorized map $\mathbf{M}$.    

However, existing methods are sensitive to observation-length mismatch, where performance degrades when the observation length is shorter than the length used during training. Our goal is therefore to design a predictor $p_\phi(\mathbf{Y}|\mathbf{X}^v,\mathbf{M})$ that remains effective with incomplete observations $\mathbf{X}^v$ and achieves results comparable to those obtained with complete observations. We define $\mathbf{X}^v \in \mathbb{R}^{N \times T_v \times C_a}$ with observation length:
\begin{equation}
    T_v = T_o - v \cdot \Delta T, 
\end{equation} 
where $v \in \{1, 2, \dots, \tau\}$ and $\tau = \frac{T_o}{\Delta T}-1$. Here, $\Delta T$ is the temporal interval omitted at each step, $\tau$ is the maximum admissible number of omissions, and $v$ indexes the number of omitted intervals. Thus, $\mathbf{X}^v$ denotes an incomplete observation in which the first $v \cdot \Delta T$ timesteps are omitted.

\begin{figure*}[!t]
    \centering
    \includegraphics[width=0.85\linewidth,trim={5pt, 2pt, 5pt, 2pt}, clip]{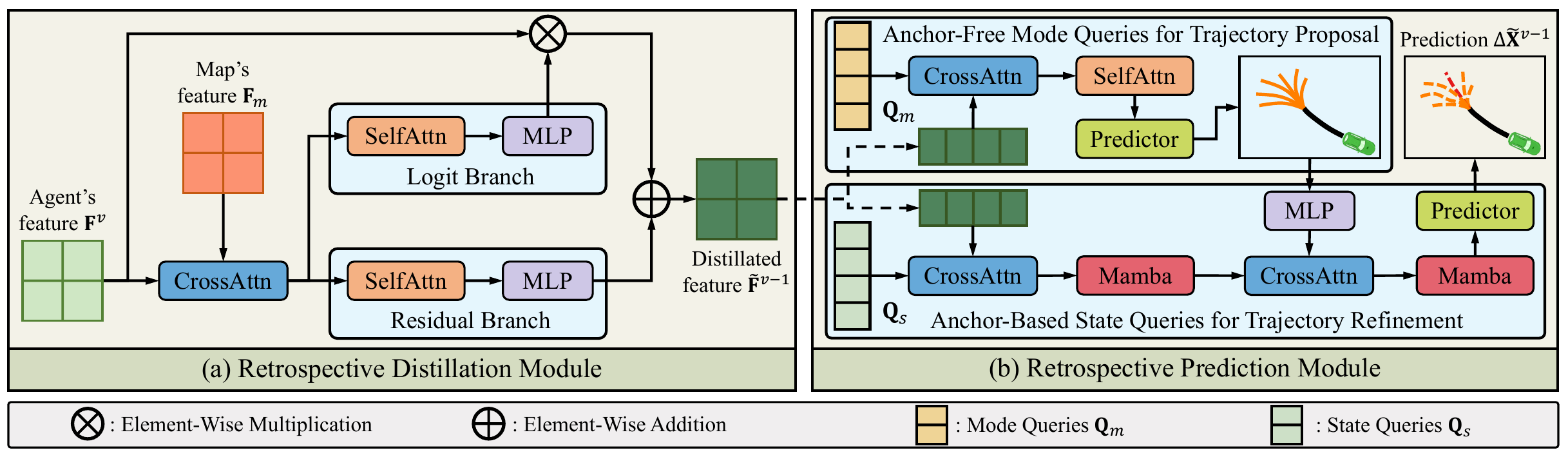}
    \vspace{-2mm}
    \caption{Illustration of the (a) RDM and (b) RPM. RDM employs a residual-based distillation strategy, featuring a logit branch that generates a gating vector and a residual branch that learns from the omitted history. RPM employs a decoupled query strategy, utilizing mode queries for multimodal trajectory proposals and state queries for trajectory refinement, with the proposals serving as anchors.}
    \label{fig:modules}
    \vspace{-5mm}
\end{figure*}

\vspace{-1mm}
\subsection{Progressive Retrospective Framework} \label{sec.3-2}
\vspace{-1mm}

Fig.~\ref{fig:teaser-c} and Fig.~\ref{fig:teaser-d} show that the performance gap narrows as the length of incomplete observations approaches the standard length. 
This can be attributed to the increased robustness of features extracted from longer observations, which motivates us to retrospect the incomplete observation to the standard length. 
However, directly recovering the omitted timesteps in a single step is challenging due to the large information gap between short and standard observations. We therefore propose a Progressive Retrospective Framework (PRF) that progressively maps incomplete trajectories to the standard length, as illustrated in Fig.~\ref{fig:arch}.

Given a dataset with standard observations $\mathbf{X}$ of length $T_o$, PRF contains $\tau$ retrospective units, each responsible for retrospecting observations of specific length to its former $\Delta T$ timesteps. 
For example, unit $\Phi^v$ reconstructs the segment $\Delta\mathbf{X}^{v-1} \in \mathbb{R}^{N_a \times \Delta T \times 2}$ between $\mathbf{X}^v$ and $\mathbf{X}^{v-1}$. The units process incomplete observations sequentially, progressively approximating the standard observations. 
Specifically, a incomplete input $\mathbf{X}^v$ is passed through $\Phi^v$, $\Phi^{v-1}$, $\dots$, and $\Phi^1$ to reconstruct the observation $\Delta\mathbf{X}^{v-1}$, $\Delta\mathbf{X}^{v-2}$, $\dots$, $\Delta\mathbf{X}^0$ until reaching the standard-length observation $\mathbf{X}$.

To make the framework plug-and-play and highly efficient, we employ a shared encoder to extract features from variable-length observations:
\begin{equation}
    \mathbf{F}^v = \operatorname{Encoder}(\mathbf{X}^v), v \in [1, 2, \dots, \tau],
\end{equation}
The unit $\Phi^v$ then takes $\mathbf{F}^v$ as input. Instead of retrospecting features or trajectories, each unit retrospects both:
\begin{equation}
    \begin{aligned}
    \tilde{\mathbf{F}}^{v-1}, \Delta\tilde{\mathbf{X}}^{v-1} &= \Phi^{v}(\mathbf{F}_v), \\
    \tilde{\mathbf{X}}^{v-1} &= \operatorname{Concat}(\Delta\tilde{\mathbf{X}}^{v-1}, \mathbf{X}^{v}),
    \end{aligned}
\end{equation}
to approximate trajectory $\mathbf{X}^{v-1}$ and its feature $\mathbf{F}^{v-1}$. Specifically, each unit comprises a Retrospective Distillation Module (RDM) and a Retrospective Prediction Module (RPM), where RDM distills features while RPM recovers omitted timesteps using the distilled features:
\begin{equation}
    \tilde{\mathbf{F}}^{v-1} = \Phi^v_D(\mathbf{F}^v),
    \quad 
    \Delta\tilde{\mathbf{X}}^{v-1} = \Phi^v_P(\tilde{\mathbf{F}}^{v-1}).
\end{equation}
At inference time, $\mathbf{F}^v$ is propagated iteratively through $v$ units to produce a standard-length feature $\tilde{\mathbf{F}^0}$, which is then passed to a shared decoder to predict future trajectory $\tilde{\mathbf{Y}} = \operatorname{Decoder}(\tilde{\mathbf{F}}^0)$.

Since a shared encoder extracts teacher and student features in RDM, feature conflict may arise during distillation. We therefore design the RDM with a residual-based distillation strategy, which models the feature of omitted $\Delta T$ time steps as learnable residuals. To further strengthen distillation, we design the RPM to recover the omitted timesteps from retrospected features, providing implicit supervision for RDM and yielding additional performance gains. These two modules enable the retrospective units to realize progressive feature distillation, significantly improving variable-length trajectory prediction.

\vspace{-1mm}
\subsection{Retrospective Distillation Module} \label{sec.3-3}
\vspace{-1mm}

Fig.~\ref{fig:modules}a illustrates the RDM. 
%
RDM models the teacher-student discrepancy induced by the omitted $\Delta T$ timesteps as residual, and adopts a residual-based distillation strategy.
RDM $\Phi^v_D$ distills student feature $\mathbf{F}^v$ of length $T_v$ to teacher feature $\mathbf{F}^{v-1}$ of length $T_{v-1}$. Since the HD map is independent of trajectory length, RDM first conditions agent features on the scene context via cross-attention:
\begin{equation}
    \mathbf{F}_m^v = \operatorname{CrossAttn}(Q=\mathbf{F}^v,K,V=\mathbf{F}_m),
\end{equation}
where $\mathbf{F}_m$ denotes the encoded feature of map $\mathbf{M}$, thereby extracting environment constraints for distillation. RDM then employs two parallel branches, a logit branch that generates element-wise gates and a residual branch that learns the residual corresponding to the omitted timesteps:
\begin{equation}
    \begin{aligned}
    \mathbf{H}_g^v &= \operatorname{SelfAttn}(Q,K,V=\mathbf{F}_m^v),\\
    \mathbf{g}^v  &= \operatorname{Sigmoid}(\operatorname{LN}(\operatorname{MLP}( [\mathbf{H}_g^v \Vert\, \mathbf{F_m}]))),\\
    \mathbf{H}_r^v &= \operatorname{SelfAttn}(Q,K,V=\mathbf{F}_m^v),\\
    \mathbf{F}^v_r &= \operatorname{ReLU}(\operatorname{LN}(\operatorname{MLP}([\mathbf{H}_r^v \Vert\, \mathbf{F_m}]))),
    \end{aligned}
\end{equation}
where $[\cdot\Vert\cdot]$ denotes concatenation, $\mathbf{g}^v$ is the gating vector, and $\mathbf{F}^v_r$ is the residual feature. Finally, the student feature is gated and fused with the learned residual feature through a shortcut connection:
\begin{equation}
    \tilde{\mathbf{F}}^{v-1} = \mathbf{g}^v \odot \mathbf{F}^v + \mathbf{F}^v_r,
\end{equation}
where $\odot$ represents element-wise multiplication. The fusion preserves reliable components through the gated shortcut, imputes omissions via the residual, and maintains gradient flow for stable, efficient training.

\vspace{-1mm}
\subsection{Retrospective Prediction Module} \label{sec.3-4}
\vspace{-1mm}

Fig.~\ref{fig:modules}b presents the RPM. 
RPM recovers the omitted $\Delta T$ timesteps from feature $\tilde{\mathbf{F}}^{v-1}$.
It adopts a decoupled query strategy to integrate anchor-free and anchor-based schemes, enabling coarse-to-fine trajectory retrospection. 
First, since retrospection is inherently multimodal, similar to prediction, RPM uses mode queries to generate diversity yet coarse multimodal proposals. Second, state queries that learns the temporal dynamics of agents treat these proposals as anchors and further refine them.

\noindent\textbf{Anchor-Free Mode Queries for Multimodal Proposal.} 
RPM employs mode queries to recover multimodal historical trajectories. Specifically, mode queries $\mathbf{Q}_m \in \mathbb{R}^{N_a \times K \times C}$, where $K$ denotes the number of motion modes, are first initialized by MLPs to preserve multimodal information. Then, cross-attention is applied to extract scene features from $\tilde{\mathbf{F}}^{v-1}$ for $\mathbf{Q}_s$. After that, self-attention is applied to $\mathbf{Q}_s$ to capture interactions among modes.
\begin{equation}
    \begin{aligned}
    \mathbf{Q}_m &= \operatorname{MLP} ([m_1, m_2, \dots, m_M]),\\
    \mathbf{Q}_m &= \operatorname{CrossAtt}(Q=\mathbf{Q}_m, K,V=\tilde{\mathbf{F}}^{v-1}),\\
    \mathbf{Q}_m &= \operatorname{SelfAttn}(Q,K,V=\mathbf{Q}_m).
    \end{aligned}
\end{equation}
Finally, a predictor composed of MLPs is used to propose multimodal trajectories using mode queries $\mathbf{Q}_m$:
\begin{equation}
    \Delta\tilde{\mathbf{X}}^{v-1}_k =\operatorname{Predictor}(\mathbf{Q}_m),
    \label{eq.9}
\end{equation}
where $\Delta\mathbf{X}^{v-1}_k  \in \mathbb{R}^{N_a \times K \times \Delta T}$ is the retrospected multimodal proposals. 

\noindent\textbf{Anchor-Based State Queries for Motion Refinement.}
RPM regards proposals $\Delta\mathbf{X}^{v-1}_k$ as anchors and utilizes state queries, which learns evolving motion dynamics, to further refine them.
Specifically, state queries $\mathbf{Q}_s \in \mathbb{R}^{N_a \times \Delta T \times C}$ are first initialized by MLPs to preserve motion dynamics. Then, cross-attention is adopted to extract scene features for $\mathbf{Q}_s$. 
After that, Mamba is conducted on $\mathbf{Q}_s$ to model agents' temporal dynamics.
\begin{equation}
    \begin{aligned}
        \mathbf{Q}_s &= \operatorname{MLP}([t_1, t_2, \dots, t_{\Delta T}]), \\
        \mathbf{Q}_s &= \operatorname{CrossAttn}(Q=\mathbf{Q}_s,K,V=\tilde{\mathbf{F}}^{v-1}),\\
        \mathbf{Q}_s &= \operatorname{Mamba}(U=\mathbf{Q}_s).
    \end{aligned}
    \label{eq.10}
\end{equation}
Next, proposals $\Delta\mathbf{X}_k^{v-1}$ are encoded to anchor features $\mathbf{F}_k^{v-1}$, on which cross-attention is performed to extract multimodal cues for $\mathbf{Q}_s$, followed by Mamba to model temporal dependencies, similar to Eq.~\ref{eq.10}. Finally, state queries that integrate multimodal property and motion dynamics are used to yield refined multimodal predictions, similar to Eq.~\ref{eq.9}. The final retrospected trajectories $\Delta\tilde{\mathbf{X}}^{v-1}$ correspond to the highest-probability mode. Given Mamba~\cite{zhu2024vision}’s strong sequence modeling capability, we employ it to model state queries over time in place of traditional attention.

Since RPM recovers a fixed $\Delta T$ timesteps independent of observation lengths, one RPM is shared across all retrospective units.
During training, with progressive distillation done upstream, distilled features are batch-processed by RPM to accelerate training. 
During inference, RPM is disabled.
Overall, RPM adds no inference cost while improving training via shared, batched supervision. 

\vspace{-1mm}
\subsection{Rolling-Start Training Strategy}
\label{sec:rsts}
\vspace{-1mm}

Existing methods use fixed $T_o$ steps to predict $T_f$ steps, so a sequence of length $T_o\!+\!T_f$ yields only one training sample, underutilizing training data. When $T_v \! < \! T_o$, the pair ($[1,T_v]$, $[T_v+1,T_v+T_f]$) forms a distinct training window, yet prior works either cannot accommodate such partial inputs or exhibit degraded performance. In contrast, PRF natively learns from shorter trajectories, enabling effective training on partial histories. We exploit this property with a Rolling-Start Training Strategy (RSTS) to improve data efficiency.

Using Argoverse 2~\cite{Argoverse2} as a concrete example. In this setup, $T_o=50$, $T_f=60$, and $\Delta T\!=\!10$. 
To train on Argoverse~2, PRF includes four retrospective units $\Phi^4$, $\Phi^3$, $\Phi^2$, $\Phi^1$, which distill features of lengths 10, 20, 30, and 40 into features of length 20, 30, 40, and 50, respectively.

RSTS begins with a standard sample ([1,50], [51,110]) pair, where observation windows \{[41,50], [31,50], [21,50], [11,50], [1,50]\} are encoded to train retrospective units \{$\Phi^{4}$, $\Phi^{3}$, $\Phi^{2}$, $\Phi^{1}$\}, with the feature of [1,50] to train decoder. The start point is then shifted to $T_v\!=\!40$, yielding a new sample pair ([1,40],[41,100]), and windows \{[31, 40], [21, 40], [11, 40], [1, 40]\} are encoded to train \{$\Phi^{4}$, $\Phi^{3}$, $\Phi^{2}$\}, with $\Phi^1$ distilling the feature of [1, 40] to standard length for decoder training. 
Similarity,  for $T_v\!=\!30$, new sample pair is used to train $\{\Phi^{4}, \Phi^{3}$\} and decoder; for $T_v\!=\!20$, new sample pair is used to train $\Phi^4$ and decoder.

As described above, a sequence yields 4 samples for decoder training and \{4, 3, 2, 1\} samples for the retrospective units $\{\Phi^4, \Phi^3, \Phi^2, \Phi^1\}$, respectively. The number of samples generated for each unit is inversely proportional to the observation length of its input. This aligns with intuition, shorter observation windows are harder to retrospect their history, and therefore benefit from more training data.

\begin{table*}[!t]
    \centering
    \resizebox{\linewidth}{!}{%
    {\setlength{\tabcolsep}{3pt}
    \begin{tabular}{l|cccccc|ccccc}
        \toprule
        \multirow{2}*{Method} & \multicolumn{6}{c|}{\textbf{Argoverse~2} (mADE$_6$/mFDE$_6$)} &
        \multicolumn{5}{c}{\textbf{Argoverse~1} (mADE$_6$/mFDE$_6$)} \\
        \cline{2-12}
        ~ & 10 & 20 & 30 & 40 & 50 & Avg-$\Delta$50 &
        5 & 10 & 15 & 20 & Avg-$\Delta$20 \\
        \midrule
        QCNet-Ori & 0.900/1.526 & 0.777/1.338 & 0.752/1.296 & 0.725/1.252 & 0.726/1.253 & 0.063/0.100 & 0.807/1.172 & 0.769/1.139 & 0.751/1.104 & 0.709/1.040 & 0.067/0.098 \\
        QCNet-IT & 0.741/1.293 & 0.734/1.279 & 0.730/1.276 & 0.726/1.267 & 0.726/1.253 & \textbf{0.007}/0.034 & 0.747/1.083 & 0.721/1.058 & 0.714/1.043 & 0.709/1.040 & 0.018/0.021 \\
        QCNet-DTO & 0.768/1.315 & 0.739/1.270 & 0.735/1.269 & 0.732/1.261 & 0.731/1.258 & 0.012/0.021 & 0.764/1.102 & 0.722/1.057 & 0.709/1.046 & 0.702/1.034 & 0.030/0.034 \\
        QCNet-FLN & 0.752/1.274 & 0.735/1.253 & 0.731/1.243 & 0.729/1.231 & 0.724/1.231 & 0.013/0.019 & 0.760/1.088 & 0.719/1.041 & 0.710/1.027 & 0.699/1.017 & 0.031/0.035 \\
        QCNet-LaKD & 0.739/1.259 & \underline{0.725}/1.235 & \underline{0.725}/1.232 & 0.721/1.227 & 0.718/1.219 & \underline{0.010}/0.019 & 0.737/1.057 & \underline{0.708}/1.044 & 0.699/1.034 & \underline{0.696}/1.027 & 0.019/0.018 \\
        QCNet-CLLS & \underline{0.735}/\underline{1.247} & 0.727/\underline{1.232} & \underline{0.725}/\underline{1.227} & \underline{0.719}/\underline{1.222} & \underline{0.714}/\underline{1.215} & 0.013/\underline{0.017} &
        \underline{0.729}/\underline{1.041} & \underline{0.708}/\underline{1.023} & \underline{0.697}/\underline{1.016} & 0.697/\underline{1.012} & \underline{0.014}/\underline{0.015} \\
        \midrule
        \rowcolor{gray!15}
        QCNet-PRF & \textbf{0.727}/\textbf{1.213} & \textbf{0.711}/\textbf{1.181} & \textbf{0.706}/\textbf{1.169} & \textbf{0.702}/\textbf{1.164} & \textbf{0.702}/\textbf{1.166} & \underline{0.010}/\textbf{0.016} & \textbf{0.699}/\textbf{1.015} & \textbf{0.686}/\textbf{0.997} & \textbf{0.677}/\textbf{0.989} & \textbf{0.675}/\textbf{0.986} &  \textbf{0.012}/\textbf{0.014}\\
        \hline\hline
        DeMo-Ori & 0.861/1.533 & 0.700/1.358 & 0.671/1.306 & 0.662/1.288 & 0.658/1.278 & 0.066/0.093 & 0.781/1.267 & 0.662/1.087 & 0.624/1.011 & 0.606/1.003 & 0.083/0.119 \\
        DeMo-IT & 0.675/1.318 & 0.661/1.296 & 0.660/1.293 & 0.659/1.287 & 0.658/1.278 & \textbf{0.006}/0.021 & 0.669/1.078 & 0.634/1.031 & 0.612/0.988 & 0.606/1.003 & 0.032/0.029 \\
        DeMo-DTO & 0.672/1.307 & 0.658/1.291 & 0.650/1.279 & 0.647/1.268 & 0.645/1.265 & 0.012/0.021 & 0.662/1.064 & 0.628/1.025 & 0.605/0.991 & 0.599/1.010 & 0.033/\underline{0.017} \\
        DeMo-FLN & 0.651/1.262 & 0.644/1.258 & 0.637/1.254 & 0.628/1.238 & 0.621/1.231 & 0.019/0.022 & 0.646/1.043 & 0.607/0.994 & 0.599/0.974 & 0.592/0.957 & 0.025/0.047 \\
        DeMo-LaKD & \underline{0.639}/1.262 & \underline{0.627}/1.251 & \underline{0.620}/1.243 & 0.617/1.236 & 0.617/1.232 & 0.009/\underline{0.016} &
        \underline{0.631}/1.008 & 0.593/0.976 & 0.584/0.933 & 0.581/0.929 & 0.022/0.043 \\
        DeMo-CLLS & 0.641/\underline{1.258} & 0.630/\underline{1.249} & 0.623/\underline{1.234} & \underline{0.614}/\underline{1.225} & \underline{0.615}/\underline{1.223} & 0.012/0.019 &
        0.634/\underline{0.998} & \underline{0.587}/\underline{0.959} & \underline{0.580}/\underline{0.919} & \underline{0.579}/\underline{0.922} & \underline{0.021}/0.037 \\
        \midrule
        \rowcolor{gray!15}
        DeMo-PRF & \textbf{0.617}/\textbf{1.183} & \textbf{0.603}/\textbf{1.155} & \textbf{0.598}/\textbf{1.143} & \textbf{0.599}/\textbf{1.145} & \textbf{0.596}/\textbf{1.142} & \underline{0.008}/\textbf{0.015} &
        \textbf{0.602}/\textbf{0.952} & \textbf{0.567}/\textbf{0.901} & \textbf{0.565}/\textbf{0.904} & \textbf{0.568}/\textbf{0.909} & \textbf{0.010}/\textbf{0.010} \\
        \bottomrule
    \end{tabular}}}
    \vspace{-2mm}
    \caption{Variable-length trajectory prediction comparison on Argoverse~2 (left) and Argoverse~1 (right) validation sets. For Argoverse~2, AVG–$\Delta$50 is the average difference between \{10, 20, 30, 40\} and 50. For Argoverse~1, AVG–$\Delta$20 is the average difference between \{5, 10, 15\} and 20. Best results are in \textbf{bold} and second best are \underline{underlined}.}
    \vspace{-2mm}
    \label{tab:variable_prediction}
\end{table*}

\begin{figure*}[!tbp] 
	\centering
	\begin{minipage}[b]{0.92\linewidth}
		\subfloat[DeMo-IT]{
			\begin{minipage}[b]{0.23\textwidth} 
                \label{fig.3a}
				\centering
				\includegraphics[width=\linewidth, trim={14pt 12pt 23pt 23pt}, clip]{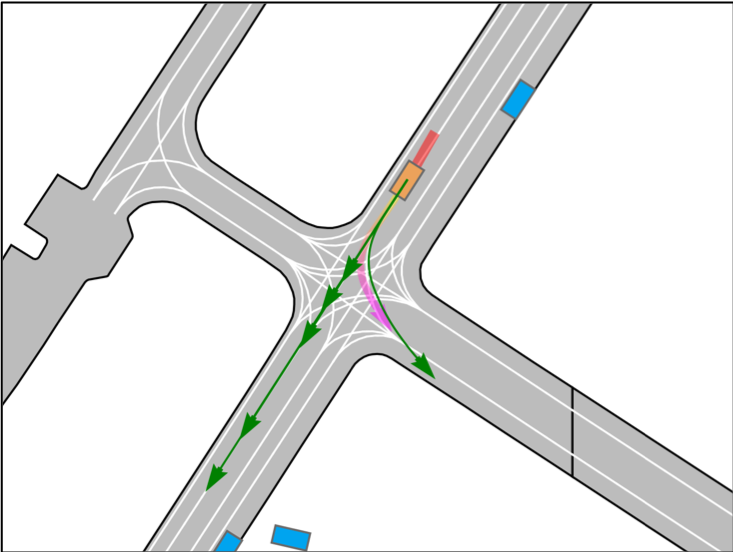}\vspace{2pt}
                \includegraphics[width=\linewidth, trim={12pt 22pt 21pt 13pt}, clip]{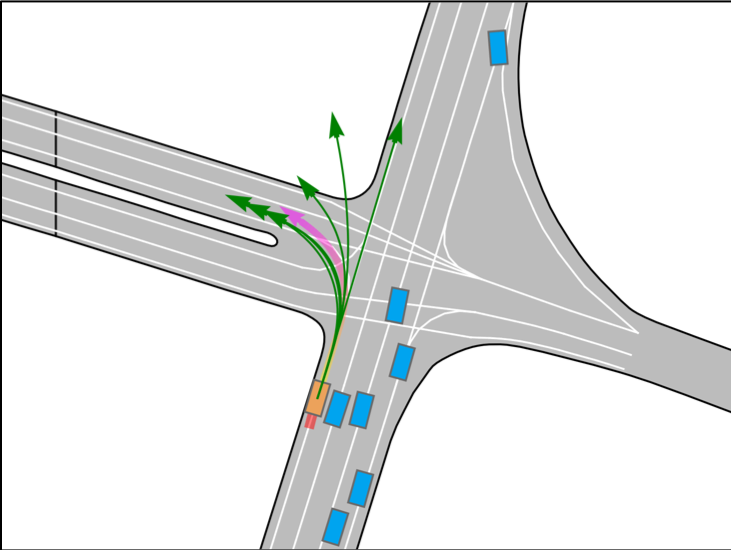}
			\end{minipage}
		}
		\hfill
		\subfloat[DeMo-CLLS]{
			\begin{minipage}[b]{0.23\textwidth}
                \label{fig.3b}
				\centering
				\includegraphics[width=\linewidth, trim={13pt 12pt 22pt 23pt}, clip]{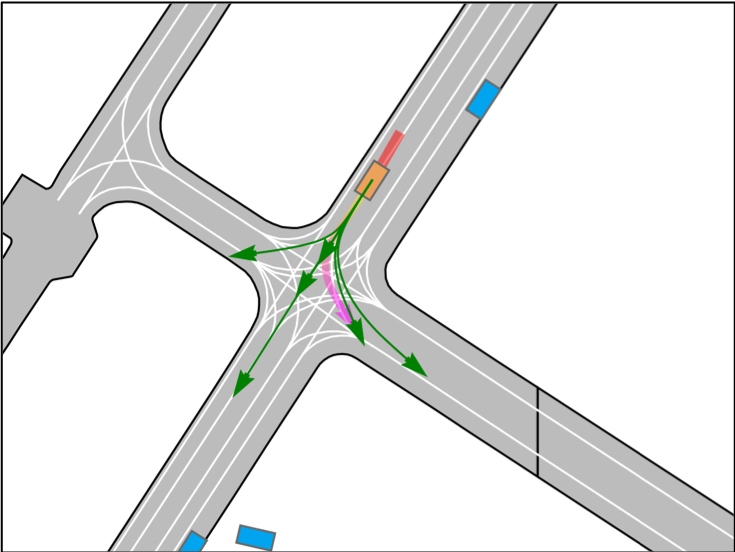}\vspace{2pt}
                \includegraphics[width=\linewidth, trim={14pt 22pt 23pt 13pt}, clip]{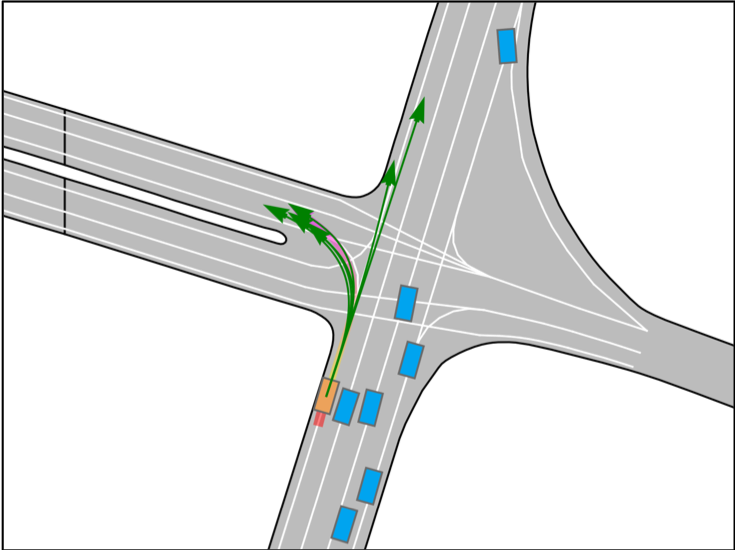}
			\end{minipage}
		}
		\hfill
		\subfloat[DeMo-PRF (Our)]{
			\begin{minipage}[b]{0.23\textwidth}
                \label{fig.3c}
				\centering
				\includegraphics[width=\linewidth, trim={13pt 12pt 22pt 23pt}, clip]{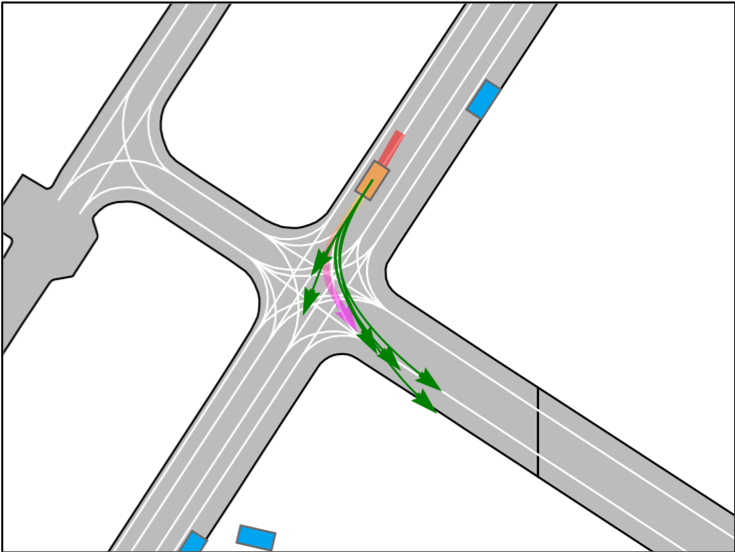}\vspace{2pt}
                \includegraphics[width=\linewidth, trim={13pt 22pt 22pt 13pt}, clip]{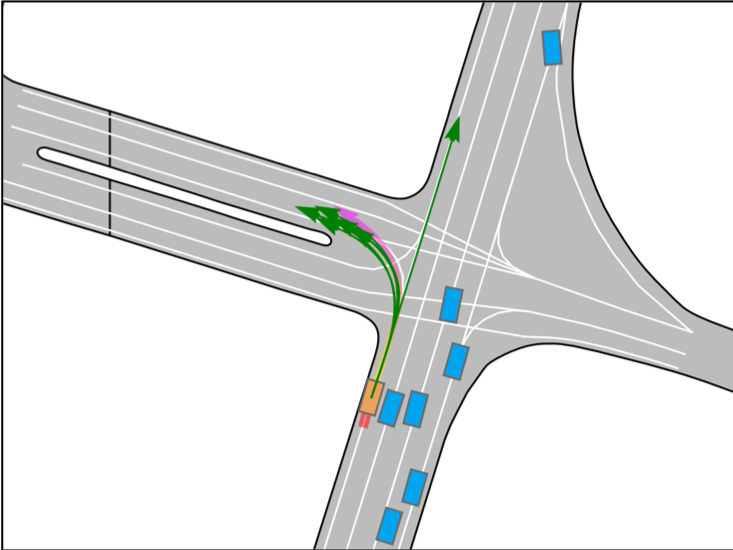}
			\end{minipage}
		}
		\hfill
		\subfloat[GT]{
			\begin{minipage}[b]{0.23\textwidth}
                \label{fig.3c}
				\centering
				\includegraphics[width=\linewidth, trim={13pt 12pt 22pt 23pt}, clip]{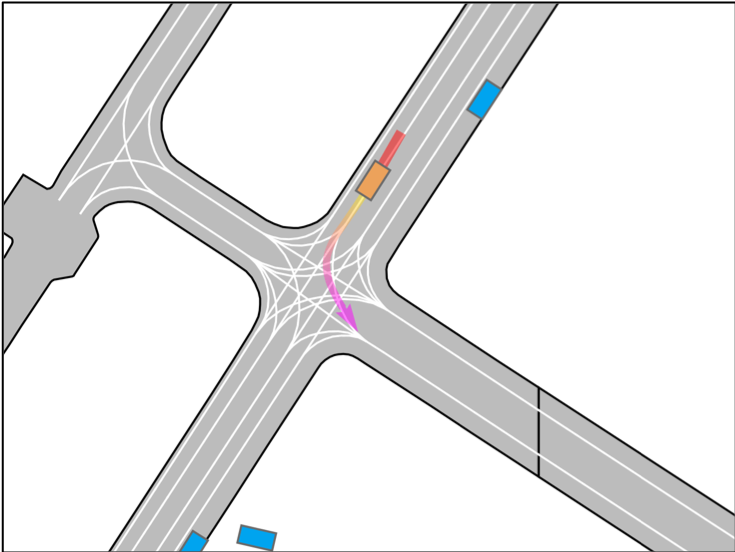}\vspace{2pt}
                \includegraphics[width=\linewidth, trim={13pt 22pt 22pt 13pt}, clip]{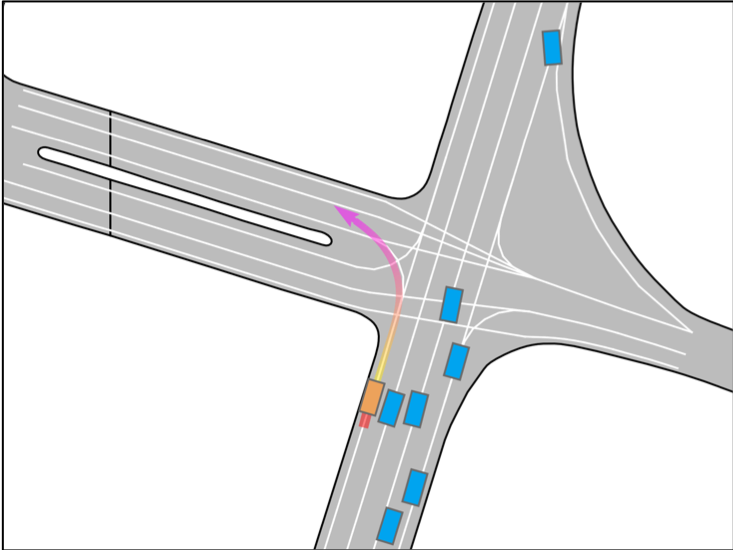}
			\end{minipage}
		}
	\end{minipage}
    \vspace{-2mm}
	\caption{Qualitative results on the Argoverse~2 validation set. Incomplete observations, predicted trajectories, and ground truth trajectories are shown in yellow, green, and pink, respectively. Our predictions align more closely with the ground truth than other methods.}
    \vspace{-5mm}
	\label{fig:results_vis}
\end{figure*}

\vspace{-1mm}
\subsection{Loss Functions}
\label{sec:}
\vspace{-1mm}

We train the decoder, RPM, and RDM end-to-end. Accordingly, the overall objective comprises three components. 
For the decoder, we adopt the same settings as QCNet~\cite{zhou2023query} and DeMo~\cite{zhang2024decoupling}, which use a smooth-L1 loss and a cross-entropy loss to supervise the trajectory regression and probability score classification, respectively.
For RPM, we adopt the same losses as the decoder, applying them twice to supervise mode queries and state queries, respectively:
\begin{equation}
    \mathcal{L}_{rpm} = \frac{1}{\tau}\sum\nolimits_{v=1}^{v=\tau} (\mathcal{L}_{mq}^v + \mathcal{L}_{sq}^v),
\end{equation}
where $\mathcal{L}_{mq}^v$ and $\mathcal{L}_{sq}^v$ are losses for mode queries and state queries in the $v$-th RPM, respectively. 
For RDM, we use a smooth-L1 loss to supervise retrospective distilling:
\begin{equation}
    \begin{aligned}
    \mathcal{L}_{dist}^v &= \operatorname{SmoothL1}(\tilde{\mathbf{F}}^{v-1}, \mathbf{F}^{v-1}),\\
    \mathcal{L}_{rdm} &= \frac{1}{\tau}\sum\nolimits_{v=1}^{v=\tau}\mathcal{L}_{dist}^v,
    \end{aligned}
\end{equation}
where $\mathcal{L}_{dist}^v$ is the distilliation loss for the $v$-th RDM. The total loss sums the losses for the decoder, RPM, and RDM.
\vspace{-2mm}
\section{Experiments}
\vspace{-1mm}
\label{sec:exp}

\subsection{Experimental Settings}
\label{sec:exp_settings}
\vspace{-1mm}

\noindent \textbf{Dataset.} We evaluate PRF on two motion forecasting datasets, Argoverse 2~\cite{Argoverse2} and Argoverse 1~\cite{Argoverse}. Argovese 2 contains 250{,}000 driving scenarios collected from six cities. Each scenario is an 11~s sequence sampled at 10~Hz, with the first 5~s as history and the subsequent 6~s forming the prediction horizon. Argoverse 1 dataset comprises 324{,}557 scenarios collected in Miami and Pittsburgh. Each scenario is a 5~s sequence sampled at 10~Hz, with the first 2~s as history and the remaining 3~s as the prediction horizon.

\noindent \textbf{Evaluation Metrics.} We use minimum Average Displacement Error (mADE$_K$), minimum Final Displacement Error (mFDE$_K$), Brier-minimum Final Displacement Error (b-mFDE$_K$), and Miss Rate (MR$_K$). 
Here, $K$ denotes the number of motion modes, and following common practice, we report results for $K\!=1\!$ and $K\!=\!6$. mADE$_k$ calculates the average error between ground-truth and prediction, while mFDE$_K$ measures the error at the endpoint. b-mFDE$_K$ extends mFDE$_K$ by incorporating the predicted mode probabilities, penalizing endpoint errors more heavily when the assigned probability is low. MR$_K$ computes the proportion of minFDE$_K$ that exceeds 2 meters.

\noindent \textbf{Backbone \& Baselines.} PRF is plug-and-play with existing prediction models. To demonstrate its compatibility, we integrate PRF with two state-of-the-arts, \textbf{QCNet}~\cite{zhou2023query} and \textbf{DeMo}~\cite{zhang2024decoupling}. To verify its effectiveness, we compare it with four closely related works, \textbf{DTO}~\cite{monti2022many}, \textbf{FLN}~\cite{xu2024adapting}, \textbf{LaKD}~\cite{li2024lakd}, and \textbf{CLLS}~\cite{qiu2025adapting}. 
We also include two baselines, \textbf{Ori} trained on standard observations and evaluated on variable-length inputs, and \textbf{IT} (Isolated Training), which trains a separate model for each observation length and evaluates on the matching length.

\begin{table*}[!t]
  \centering
  \begin{minipage}[t]{0.529\textwidth}
    \centering
    \resizebox{\linewidth}{!}{%
      \begin{tabular}{l|cccccc}
        \toprule
        Method & b-mFDE$_6$ & mADE$_6$ & mFDE$_6$ & MR$_6$ & mADE$_1$ & mFDE$_1$\\
        \midrule
        FRM~\cite{park2023leveraging} & 2.47 & 0.89 & 1.81 & 0.29 & 2.37 & 5.93\\
        HDGT~\cite{jia2023hdgt} & 2.24 & 0.84 & 1.60 & 0.21 & 2.08 & 5.37\\
        THOMAS~\cite{gilles2022thomas} & 2.16 & 0.88 & 1.51 & 0.20 & 1.95 & 4.71\\
        SIMPL~\cite{zhang2024simpl} & 2.05 & 0.72 & 1.43 & 0.19 & 2.03 & 5.50\\
        HPTR~\cite{zhang2023real} & 2.03 & 0.73 & 1.43 & 0.19 & 1.84 & 4.61\\
        GoRela~\cite{cui2023gorela} & 2.01 & 0.76 & 1.48 & 0.22 & 1.82 & 4.62\\
        MTR~\cite{shi2022motion} & 1.98 & 0.73 & 1.44 & \underline{0.15} & 1.74 & 4.39\\
        GANet~\cite{wang2023ganet} & 1.96 & 0.72 & 1.34 & 0.17 & 1.77 & 4.48\\
        DeMo~\cite{zhang2024decoupling} & 1.92 & 0.65 & 1.25 & 0.15 & \underline{1.58} & 3.96\\
        QCNet~\cite{zhou2023query} & 1.91 & 0.65 & 1.29 & 0.16 & 1.69 & 4.30\\
        ReMo~\cite{song2024motion} & 1.89 & 0.66 & 1.24 & \underline{0.15} & 1.59 & 3.93\\
        Tamba~\cite{huang2025trajectory} & 1.89 & 0.64 & 1.24 & 0.17 & 1.66 & 4.24\\
        ProphNet~\cite{wang2023prophnet} & 1.88 & 0.68 & 1.33 & 0.18 & 1.80 & 4.74\\
        SmartRefine~\cite{zhou2024smartrefine} & 1.86 & 0.63 & 1.23 & \underline{0.15} & 1.65 & 4.17\\
        DeMo+ReMo~\cite{zhang2024decoupling,song2024motion} & \underline{1.84} & \underline{0.61} & \underline{1.17} & \textbf{0.13} & \textbf{1.49} & \underline{3.74}\\
        \midrule
        \rowcolor{gray!15}
        DeMo-PRF~(Our) & \textbf{1.81} & \textbf{0.60} & \textbf{1.14} & \textbf{0.13} & \textbf{1.49} & \textbf{3.72}\\
        \bottomrule
      \end{tabular}%
    }
    \vspace{-2mm}
    \captionof{table}{Comparison with state-of-the-arts on the Argoverse~2 Single Agent Motion Forecasting Leaderboard ranked by b-mFDE$_6$. All results are from a single model, without model ensembling.}
    \label{tab:arg2_standard}
  \end{minipage}\hfill
  \begin{minipage}[t]{0.40\textwidth}
    \centering
    \resizebox{\linewidth}{!}{%
      \begin{tabular}{l|cccc}
        \toprule
        Method & b-mFDE$_6$ & mADE$_6$ & mFDE$_6$ & MR$_6$ \\
        \midrule
        LaneGCN~\cite{liang2020learning} & 2.06 & 0.87 & 1.36 & 0.16\\
        mmTrans.~\cite{liu2021multimodal} & 2.03 & 0.84 & 1.34 & 0.15\\
        DenseTNT~\cite{gu2021densetnt} & 1.98 & 0.88 & 1.28 & 0.13\\
        TPCN~\cite{ye2021tpcn} & 1.93 & 0.82 & 1.24 & 0.13\\
        SceneTrans.~\cite{ngiamscene} & 1.89 & 0.80 & 1.23 & 0.13\\
        HOME+GOME~\cite{gilles2021home,gilles2022gohome} & 1.86 & 0.89 & 1.29 & \textbf{0.08}\\
        HiVT~\cite{zhou2022hivt} & 1.84 & 0.77 & 1.17 & 0.13\\
        MultiPath++~\cite{varadarajan2022multipath++} & 1.79 & 0.79 & 1.21 & 0.13\\
        GANet~\cite{wang2023ganet} & 1.79 & 0.81 & 1.16 & 0.12\\
        PAGA~\cite{da2022path} & 1.76 & 0.80 & 1.21 & 0.11\\
        MISC~\cite{ye2022dcms} & 1.76 & 0.77 & 1.14 & 0.11\\
        Wayformer~\cite{nayakanti2023wayformer} & 1.74 & 0.77 & 1.16 & 0.12\\
        HPNet~\cite{tang2024hpnet} & 1.74 & 0.76 & 1.10 & 0.11\\
        QCNet~\cite{zhou2023query} & \underline{1.69} & 0.73 & \underline{1.07} & 0.11\\
        Tamba~\cite{huang2025trajectory} & \textbf{1.67} & \underline{0.72} & \underline{1.07} & \underline{0.09}\\
        \midrule
        \rowcolor{gray!15}
        DeMo-PRF~(Our) & 1.73 & \textbf{0.70} & \textbf{1.03} & 0.11\\
        \bottomrule
      \end{tabular}%
    }
    \vspace{-2mm}
    \captionof{table}{Comparison with state-of-the-arts on the Argoverse~1 Motion Forecasting Leaderboard. All results are from a single model, without model ensembling.}
    \label{tab:arg1_standard}
  \end{minipage}
  \vspace{-4mm}
\end{table*}

\begin{table}[!t]
    \centering
    \resizebox{\linewidth}{!}{%
    \setlength{\tabcolsep}{3pt}
    \begin{tabular}{ccc|ccccc}
        \toprule
        \multirow{2}*{RDM} & \multirow{2}*{RPM} & \multirow{2}*{RSTS} & \multicolumn{5}{c}{mADE$_6$/mFDE$_6$}\\
        \cline{4-8}
        ~ & ~ & ~ & 10 & 20 & 30 & 40 & 50\\
        \midrule
         &  &  & 0.876/1.455 & 0.769/1.337 & 0.756/1.286 & 0.726/1.252 & 0.725/1.256\\
        \checkmark &  &  & 0.655/1.257 & 0.640/1.237 & 0.636/1.231 & 0.636/1.227 & 0.639/1.231\\
        \checkmark & \checkmark &  & 0.652/1.241 & 0.637/1.214 & 0.634/1.207 & 0.631/1.204 & 0.635/1.208\\
        \rowcolor{gray!15}
        \checkmark & \checkmark & \checkmark & \textbf{0.617}/\textbf{1.183} & \textbf{0.603}/\textbf{1.155} & \textbf{0.598}/\textbf{1.143} & \textbf{0.599}/\textbf{1.145} & \textbf{0.596}/\textbf{1.142}\\
        \bottomrule
    \end{tabular}}
    \vspace{-2mm}
    \caption{Ablation study of the core modules of our model on the Argoverse~2 validation set.}
    \label{tab:ablation_modules}
    \vspace{-5mm}
\end{table}

\noindent \textbf{Implementation Details.} We define different observation lengths (timesteps) for each dataset. For Argoverse~2, with an observation horizon $T_o=50$ and a prediction horizon $T_f=60$, we set the omission interval to $\Delta T=10$, yielding variable observation lengths $\{10, 20, 30, 40, 50\}$. 
For Argoverse~1, with $T_o=20$ and $T_f=30$, we set $\Delta T =5$, yielding $\{5, 10, 15, 20\}$. 
In practice, if an observation length falls outside these sets, we truncate the observed trajectory to the nearest shorter admissible length (\textit{e.g.}, 32$\!\to\!$30), retaining the most recent timesteps. 
Models are trained end-to-end for 60 epochs with a batch size of 16, using the Adam optimizer with an initial learning rate of 0.003 and a weight decay of 0.01. 
All experiments are implemented in PyTorch and run on 8 Nvidia RTX 4090 GPUs.

\begin{table*}[!t]
\centering
    \begin{minipage}[t]{0.485\linewidth}
    \centering
    \resizebox{0.9\linewidth}{!}{%
    \begin{tabular}{@{}c|ccccc@{}}
        \toprule
        \multirow{2}{*}{Num} & \multicolumn{5}{c}{mADE$_6$/mFDE$_6$} \\
        \cmidrule(lr){2-6}
        ~ & 10 & 20 & 30 & 40 & 50 \\
        \midrule
        1 & 0.661/1.280 & 0.646/1.256 & 0.642/1.249 & 0.641/1.248 & 0.642/1.243 \\
        2 & 0.660/1.275 & 0.645/1.249 & 0.641/1.244 & 0.640/1.239 & 0.645/1.244 \\
        \rowcolor{gray!15}
        3 & \textbf{0.655}/\textbf{1.257} & \textbf{0.640}/\textbf{1.237} & \textbf{0.636}/\textbf{1.231} & \textbf{0.636}/\textbf{1.227} & \textbf{0.639}/\textbf{1.231} \\
        \bottomrule
    \end{tabular}}
    \vspace{-2mm}
    \captionof{table}{Ablation study of the number of self- and cross-attention layers in RDM on the Argoverse~2 validation set.}
    \label{tab:ablation_layer_RDM}
    \end{minipage}
    \hfill\hspace{1.5mm}
    \begin{minipage}[t]{0.485\linewidth}
    \centering
    \resizebox{0.9\linewidth}{!}{%
    \begin{tabular}{@{}c|ccccc@{}}
        \toprule
        \multirow{2}{*}{Num} & \multicolumn{5}{c}{mADE$_6$/mFDE$_6$} \\
        \cmidrule(lr){2-6}
        ~ & 10 & 20 & 30 & 40 & 50 \\
        \midrule
        1 & \textbf{0.647}/1.252 & \textbf{0.632}/1.227 & \textbf{0.628}/1.214 & \textbf{0.627}/1.211 & \textbf{0.631}/1.226 \\
        2 & 0.652/1.257 & 0.637/1.230 & 0.633/1.220 & 0.632/1.223 & 0.634/1.224 \\
        \rowcolor{gray!15}
        3 & 0.652/\textbf{1.241} & 0.637/\textbf{1.214} & 0.634/\textbf{1.207} & 0.631/\textbf{1.204} & 0.635/\textbf{1.208} \\
        \bottomrule
    \end{tabular}}
    \vspace{-2mm}
    \captionof{table}{Ablation of the number of self-, cross-attention, and Mamba layers in RPM on the Argoverse~2 validation set.}
    \label{tab:ablation_layer_RPM}
    \end{minipage}
\vspace{-4mm}
\end{table*}

\begin{table}[!t]
    \centering
    \resizebox{0.9\linewidth}{!}{%
    \begin{tabular}{c|ccccc}
        \toprule
        \multirow{2}*{Num} & \multicolumn{5}{c}{mADE$_6$/mFDE$_6$} \\
        \cline{2-6}
        ~ & 10 & 20 & 30 & 40 & 50\\
        \midrule
        GRU & 0.662/1.286 & 0.644/1.256 & 0.640/1.245 & 0.639/1.242 & 0.640/1.245\\
        Attn & 0.653/1.261 & 0.639/1.235 & 0.635/1.229 & 0.634/1.225 & \textbf{0.634}/1.224\\
        \rowcolor{gray!15}
        Mamba & \textbf{0.652}/\textbf{1.241} & \textbf{0.637}/\textbf{1.214} & \textbf{0.634}/\textbf{1.207} & \textbf{0.631}/\textbf{1.204} & 0.635/\textbf{1.208}\\
        \bottomrule
    \end{tabular}}
    \vspace{-2mm}
    \caption{Ablation of the sequence modeling choices in RPM on the Argoverse~2 validation set.}
    \vspace{-2mm}
    \label{tab:ablation_mamba}
\end{table}

\vspace{-1mm}
\subsection{Comparison with State-of-the-art} 
\vspace{-1mm}

\textbf{Variable-Length Trajectory Prediction.} The results of variable-length prediction on Argoverse~2 and Argoverse~1 validation sets are reported in Tab.~\ref{tab:variable_prediction}. Results show that PRF significantly outperforms Ori across all observation lengths, indicating the necessity of designing a framework for variable-length prediction. Secondly, IT shows modest improvement over Ori across variable-length observations, which verifies that length-specific training is expensive and brings only marginal gains. 
Moreover, PRF outperforms baselines DTO, FLN, LaKD, and CLLS across observation lengths and achieves a small performance gap between incomplete and standard observations, demonstrating state-of-the-art performance. Finally, PRF achieves the best results with both QCNet and DeMo backbones, validating its compatibility.

To qualitatively assess the superiority of PRF, we visualize results for IT, the second-best CLLS, and PRF at the shortest observation length of 10 on the Argoverse~2 validation set, as shown in Fig.~\ref{fig:results_vis}. 
The two samples respectively present complex intersection and T-junction scenarios where the agent is about to turn. The visualization shows that PRF is accurate and closer to the ground truth compared to other methods on backbone DeMo.

\noindent \textbf{Standard Trajectory Prediction.} 
%
PRF can also be extended to the standard trajectory prediction with a complete observation setting.
As shown in Tab.~\ref{tab:variable_prediction}, we further compare PRF that uses DeMo as backbone with state-of-the-art methods on the Argoverse~2 and Argoverse~1 motion forecasting benchmarks in the single agent setting with standard-length inputs, as presented in Tab.~\ref{tab:arg2_standard} and Tab.~\ref{tab:arg1_standard}. 
Tab.~\ref{tab:arg2_standard} shows that PRF achieves the best performance across all metrics on the Argoverse~2 test set, while Tab.~\ref{tab:arg1_standard} shows that PRF achieves the best performance among metrics mADE$_6$ and mFDE$_6$ on the Argoverse~1 test set. 
These results validate the generalization of PRF.

\vspace{-1mm}
\subsection{Ablation Studies}
\vspace{-1mm}

\noindent \textbf{Effects of modules.} Tab.~\ref{tab:ablation_modules} reports ablations of the core modules of PRF. The first shows the results of the DeMo backbone. 
The second row adds RDM to distill features. 
This yields substantial gains across all observation lengths, demonstrating its effectiveness. 
%
The third row further incorporates RPM to recover omitted historical trajectories.
This provides implicit supervision for distillation and delivers additional improvements at all observation lengths. 
The last row applies the RSTS. The results show consistent gains across various observation lengths, indicating that the proposed training regime enhances data utilization.

\noindent \textbf{Effects of attention layers in RDM.} Self- and cross-attention are key to retrospective distillation in RDM. We ablate the number of self- and cross-attention layers in RDM, as shown in Tab.~\ref{tab:ablation_layer_RDM}. The results show that increasing the number of layers steadily improves performance across different observation lengths. We therefore set three layers of self- and cross-attention as the default in RDM.

\noindent \textbf{Effects of attention and Mamba layers in RPM.} Self-, cross-attention, and Mamba are used in RPM to extract embedding for retrospective prediction. We ablate the number of layers for these components in RPM, as shown in Tab.~\ref{tab:ablation_layer_RPM}. The results show that a single layer yields the best mADE$_6$, while three layers yield the best mFDE$_6$. Since the gains in mFDE$_6$ are larger than those in mADE$_6$, we set three layers of self-, cross-attention, and Mamba as the default in RPM.

\begin{table}[!t]
    \centering
    \resizebox{\linewidth}{!}{%
    \setlength{\tabcolsep}{3pt}
    \begin{tabular}{ccc|ccccc}
        \toprule
        \multirow{2}*{[0,40]} & \multirow{2}*{[0,30]} & \multirow{2}*{[0,20]} & \multicolumn{5}{c}{mADE$_6$/mFDE$_6$}\\
        \cline{4-8}
        ~ & ~ & ~ & 10 & 20 & 30 & 40 & 50\\
        \midrule
         &  &  & 0.652/1.241 & 0.637/1.214 & 0.634/1.207 & 0.631/1.204 & 0.635/1.208\\
        \checkmark &  &  & 0.631/1.211 & 0.618/1.189 & 0.613/1.178 & 0.615/1.185 & 0.613/1.182\\
        \checkmark & \checkmark &  & 0.624/1.201 & 0.608/1.174 & 0.606/1.167 & 0.606/1.167 & 0.606/1.165\\
        \rowcolor{gray!15}
        \checkmark & \checkmark & \checkmark & \textbf{0.617}/\textbf{1.183} & \textbf{0.603}/\textbf{1.155} & \textbf{0.598}/\textbf{1.143} & \textbf{0.599}/\textbf{1.145} & \textbf{0.596}/\textbf{1.142}\\
        \bottomrule
    \end{tabular}}
    \vspace{-2mm}
    \caption{Ablation study of data utilization in the RSTS on the Argoverse~2 validation set.}
    \vspace{-3mm}
    \label{tab:ablation_data_RST}
\end{table}

\noindent \textbf{Effect of sequence modeling in RPM.} 
Mamba is employed to model state queries over time in RPM. To assess its effectiveness, we compare it with other modules, including GRU and Attention, as shown in Tab.~\ref{tab:ablation_mamba}. Mamba achieves the best results among these variants, confirming its superior ability to capture temporal dependencies in state queries.

\noindent \textbf{Effects of data utilization in RSTS.} RSTS improves data utilization. For example, a standard Argoverse~2 sequence with an observation length of 50 can generate additional samples with observation windows \{[0, 40], [0, 30], [0, 20]\}. We ablate these extra training samples, as shown in Tab.~\ref{tab:ablation_data_RST}. The first row reports training with only standard observations. The second through fourth rows gradually add observation windows [0, 40], [0, 30], and [0, 20] for training. The results show that incorporating incomplete observations yields steady gains in trajectory prediction, indicating that the RSTS improves data utilization.

\vspace{-1mm}
\subsection{Analysis of Distillation Strategy}
\vspace{-1mm}

PRF adopts a progressive strategy to distill features from incomplete to complete observations. 
Existing methods typically use a direct one-shot distillation strategy.
We compare our progressive distillation with this strategy by modifying PRF to directly distill features from lengths \{10, 20, 30, 40\} to length 50. As shown in Tab.~\ref{tab:analysis_direct}, our strategy outperforms direct distillation across all observation lengths, with larger gains for shorter observations. These findings indicate that progressive distillation reduces task difficulty and improves variable-length trajectory prediction.

\begin{table}[!t]
    \centering
    \resizebox{0.95\linewidth}{!}{
    \begin{tabular}{c|ccccc}
        \toprule
        \multirow{2}*{Strategy} & \multicolumn{5}{c}{mADE$_6$/mFDE$_6$} \\
        \cline{2-6}
        ~ & 10 & 20 & 30 & 40 & 50 \\
        \midrule
        Direct & 0.663/1.275 & 0.644/1.240 & 0.639/1.228 & 0.635/1.220 & 0.635/1.222\\
        \rowcolor{gray!15}
        PRF~(Our) & \textbf{0.652}/\textbf{1.241} & \textbf{0.637}/\textbf{1.214} & \textbf{0.634}/\textbf{1.207} & \textbf{0.631}/\textbf{1.204} & 0.635/\textbf{1.208} \\
        \bottomrule
    \end{tabular}}
    \vspace{-2mm}
    \caption{Analysis of progressive distillation vs. direct distillation on the Argoverse~2 validation set. RTST is not used in training.}
    \vspace{-5mm}
    \label{tab:analysis_direct}
\end{table}

\begin{figure}[!t]
  \centering
  \begin{subfigure}{0.45\linewidth}
    \centering
    \includegraphics[width=\linewidth, trim={15pt 12pt 10pt 10pt}, clip]{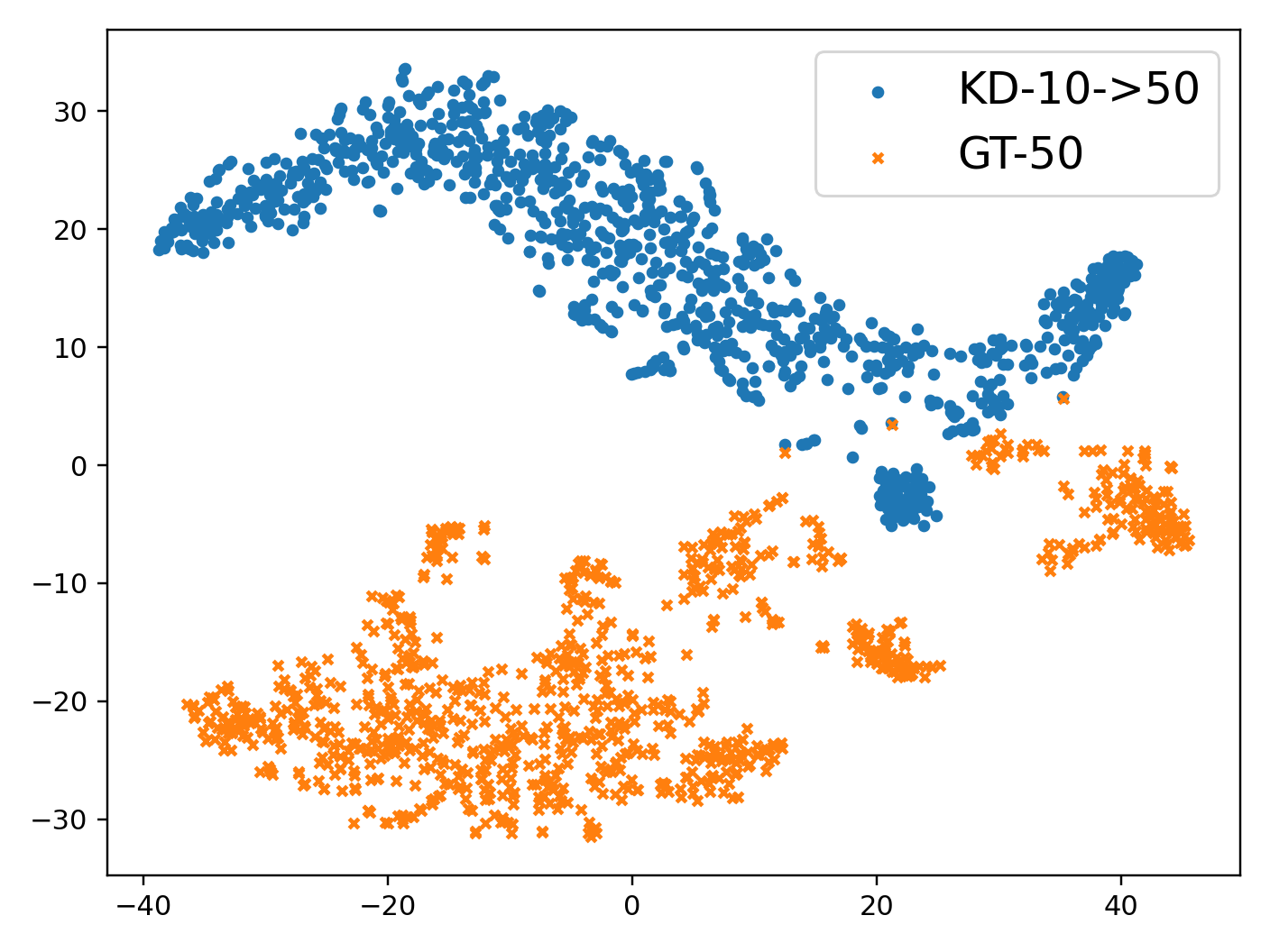}
    \caption{Direct distillation}
    \label{fig:dd}
  \end{subfigure}
  \begin{subfigure}{0.45\linewidth}
    \centering
    \includegraphics[width=\linewidth, trim={15pt 12pt 10pt 10pt}, clip]{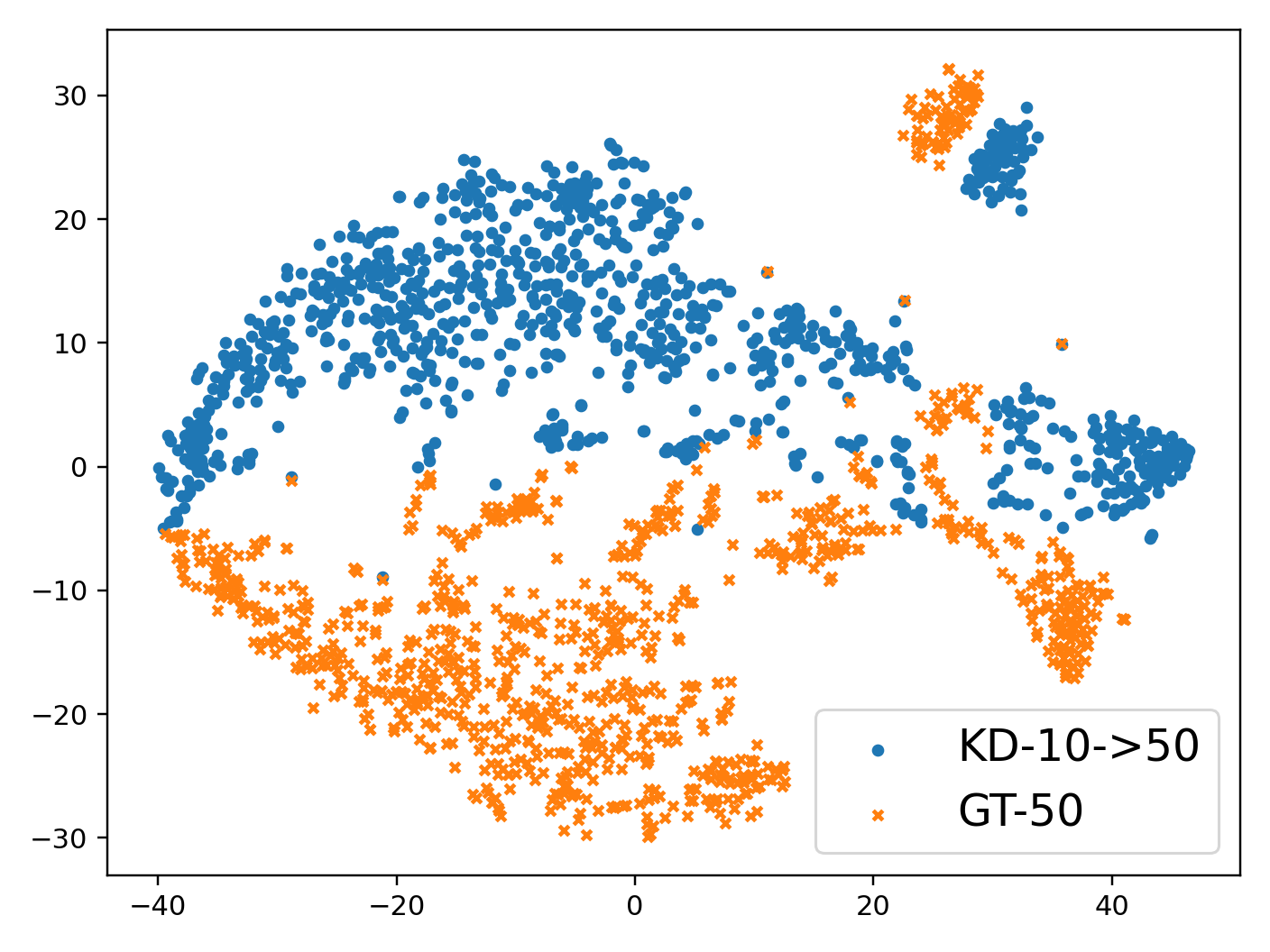}
    \caption{Progressive distillation}
    \label{fig:pd}
  \end{subfigure}
  \vspace{-3mm}
  \caption{t-SNE visualization of (a) direct and (b) progressive distillation strategies. Features distilled from 10 to 50 are shown in yellow, while features of the standard length 50 are shown in blue.}
  \vspace{-2mm}
  \label{fig:tsne-comparison}
\end{figure}

We also visualize the two strategies via t-SNE by embedding the distilled (10$\!\rightarrow$\!50) features together with the native 50-step features. Fig.~\ref{fig:dd} reveals that aligning 10-step features to 50-step ones is nontrivial, and direct distillation struggles to bridge this gap. By contrast, Fig.~\ref{fig:pd} shows that our progressive strategy achieves substantially better alignment, providing additional evidence of its effectiveness.
\begin{table}[!t]
    \centering
    \resizebox{0.8\linewidth}{!}{
    \begin{tabular}{c|ccccc}
        \toprule
         
        Length & 10 & 20 & 30 & 40 & 50 \\
        \midrule
        Inference time (s) & 0.268 & 0.236 & 0.203 & 0.172 & 0.140\\
        FLOPs (G) & 1.651 & 1.581 & 1.513 & 1.443 & 1.375\\
        \bottomrule
    \end{tabular}}
    \vspace{-2mm}
    \caption{Analysis of inference efficiency on the Argoverse~2 validation set. Results are measured with one multi-agent scenario per forward pass, using an NVIDIA GeForce RTX 4090 GPU.}
    \vspace{-5mm}
    \label{tab:model_efficiency}
\end{table}

\vspace{-1mm}
\subsection{Inference Efficiency Analysis}
\vspace{-1mm}

PRF introduces extra inference overhead by iteratively retrospecting features. We evaluate its inference cost, as shown in Tab.~\ref{tab:model_efficiency}. The cost increases almost linearly with the number of retrospective stages as the observation length decreases. 
Relative to the standard length of 50, each additional retrospective stage adds about 0.07~G FLOPs and 0.03~s of latency. 
This indicates that PRF improves prediction for incomplete lengths while incurring only a modest inference cost. 
PRF remains efficient because RDM and RST are used only during training to provide extra supervision and data, thereby incurring no test-time computation.
\vspace{-2mm}
\section{Conclusion}
\vspace{-1mm}

This paper presents a Progressive Retrospective Framework (PRF) for variable-length trajectory prediction. PRF consists of a cascade of retrospective units that progressively map incomplete-length observations to a standard length. Each unit includes a Retrospective Distillation Module (RDM) and a Retrospective Prediction Module (RFM). RDM distills features from shorter observations to align them with those from longer ones, and RPM recovers the omitted historical trajectories from the distilled features. 
To better exploit training data with incomplete observations, we further propose a Rolling-Start Training Strategy (RSTS) to improve data utilization. Extensive experiments on Argoverse~2 and Argoverse~1 demonstrate that PRF achieves state-of-the-art performance for variable-length trajectory prediction. PRF also achieves leading results for standard trajectory prediction on the Argoverse~2 and Argoverse~1 motion forecasting leaderboards.

\section{Acknowledgement}

This work was supported in part by the National Natural Science Foundation of China under Grant 62503084 and Grant 62202308, in part by the Guangdong Basic and Applied Basic Research Foundation under Grant 2024A1515110124, and in part by the Science and Technology Commission of Shanghai Municipality under Grant 24ZR1400400.
{
    \small
    \bibliographystyle{ieeenat_fullname}
    \bibliography{main}
}
\clearpage
\setcounter{page}{1}
\maketitlesupplementary

In this supplementary file, we provide additional details and results to demonstrate the benefits of the proposed framework further. The contents include the following appendices:
\begin{itemize}
    \item Appendix for RSTS (Section~\ref{sec:rsts_ape})
    \item Appendix for Loss Functions (Section~\ref{sec:losses_ape})
    \item Appendix for Evaluation Metrics (Section~\ref{sec:metrics_ape})
    \item Appendix for Qualitative Evaluations. (Section~\ref{sec:qualit_ape})
    \item Appendix for Interpretability Analysis. (Section~\ref{sec:inter_ape})
\end{itemize}

\vspace{-1mm}
\section{Appendix for RSTS}
\vspace{-1mm}
\label{sec:rsts_ape}

\begin{figure}
    \centering
    \includegraphics[width=\linewidth, trim={10pt 5pt 10pt 5pt},clip]{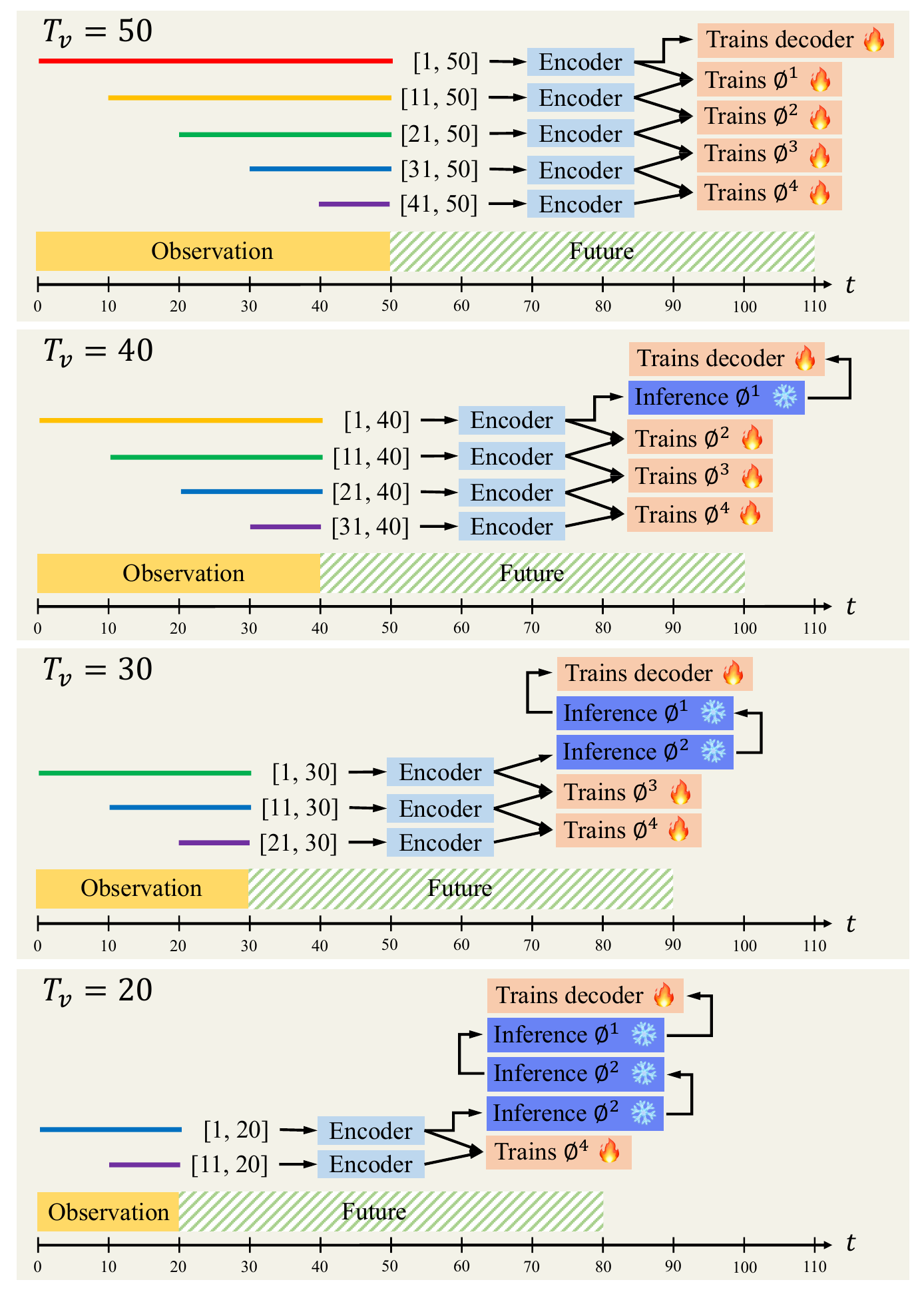}
    \vspace{-2mm}
    \caption{Illustration of the RSTS on the Argoverse~2 dataset with a standard observation horizon of $T_o=50$ and a prediction horizon of $T_f=60$. As the prediction start point shifts from 50 to 40, 30, and 20, additional training samples are generated to train the retrospective units and the decoder.}
    \vspace{-5mm}
    \label{fig:rsts_ape}
\end{figure}

The proposed Rolling-Start Training Strategy (RSTS), described in Section~\ref{sec:rsts}, improves data efficiency by incorporating incomplete observations during training. Fig.~\ref{fig:rsts_ape} illustrates the applications of RSTS on the Argoverse~2 dataset, with a standard observation horizon of $T_o=50$ and a prediction horizon of $T_f=60$.

When $T_v=50$, which corresponds to the standard observation horizon, a standard sample pair ([1,50], [51,110]) can be segmented into observation windows \{[41,50], [31,50], [21,50], [11,50], [1,50]\}. These observation windows are then encoded to train retrospective units \{$\Phi^{4}$, $\Phi^{3}$, $\Phi^{2}$, $\Phi^{1}$\}, with the encoded feature of the standard-length observation window [1,50] being used to train the decoder. 

Then, the start point is shifted to $T_v=40$, generating a sample pair ([1,40],[41,100]). This sample pair is segmented into observation windows \{[31, 40], [21, 40], [11, 40], [1, 40]\}. These observation windows are encoded to train respective units \{$\Phi^{4}$, $\Phi^{3}$, $\Phi^{2}$\}. The encoded feature of the incomplete observation window [1, 40] is distilled by unit $\Phi^{1}$ to match the standard observation length, which is then used to train the decoder.

Subsequently, the start point is shifted to $T_v=30$, producing a sample pair ([1, 30],[31,90]). This sample pair is segmented into observation windows \{[21, 30], [11, 30], [1, 30]\}. These observation windows are encoded to train respective units \{$\Phi^{4}$, $\Phi^{3}$\}. The encoded feature of the incomplete observation window [1,30] is sequentially distilled by units $\Phi^{2}$ and $\Phi^{1}$ to match the standard observation length, which is used to train the encoder.

Finally, the start point is shifted to $T_v=20$, yielding a sample pair ([1,20],[21,80]). This sample pair is segmented into observation windows \{[11, 20], [1, 20]\}. The two observation windows are encoded to train unit $\Phi^{4}$, with the encoded feature of the incomplete observation window [1, 20] being sequentially distilled by units $\Phi^{3}$, $\Phi^{2}$, and $\Phi^{1}$ to match the standard observation length, which is used to train the decoder.

In summary, RSTS generates \{4,3,2,1\} samples to train the retrospective units \{$\Phi^{4}$, $\Phi^{3}$, $\Phi^{2}$, $\Phi^{1}$\}, respectively, and 4 samples to train the decoder, using a standard training sequence.

\begin{figure*}[!t] 
	\centering
	\begin{minipage}[b]{0.81\linewidth}
		\subfloat[DeMo-IT]{
			\begin{minipage}[b]{0.225\textwidth} 
                \label{fig.6a}
				\centering
                \fbox{
                \includegraphics[width=\linewidth, trim={13pt 12pt 22pt 23pt}, clip]{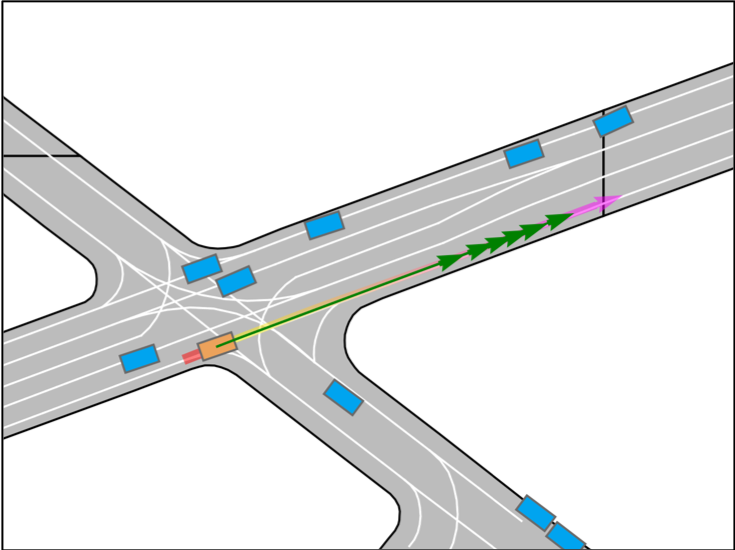}}\vspace{2pt}
                \fbox{
                \includegraphics[width=\linewidth, trim={13pt 12pt 12pt 15pt}, clip]{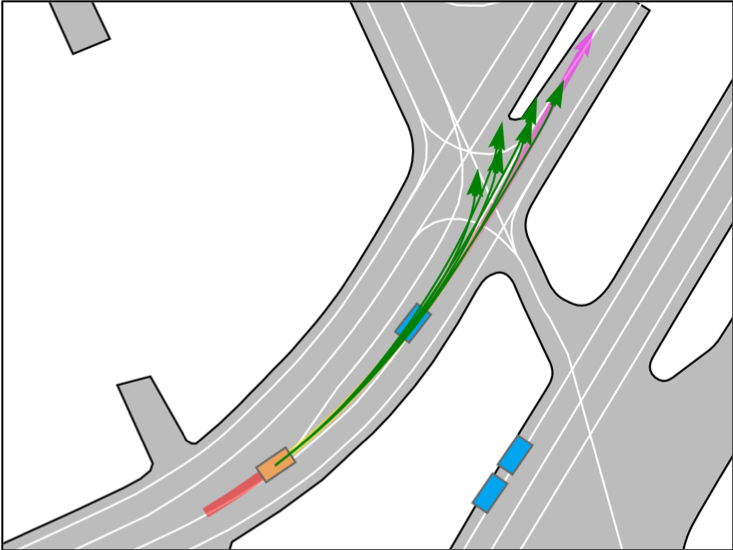}}\vspace{2pt}
                \fbox{
				\includegraphics[width=\linewidth, trim={13pt 12pt 22pt 23pt}, clip]{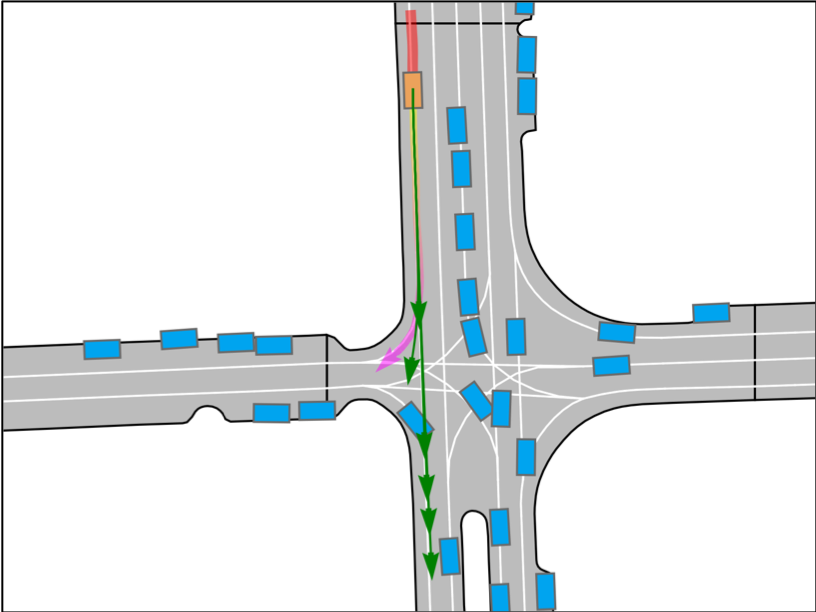}}\vspace{2pt}
                \fbox{
                \includegraphics[width=\linewidth, trim={14pt 12pt 22pt 23pt}, clip]{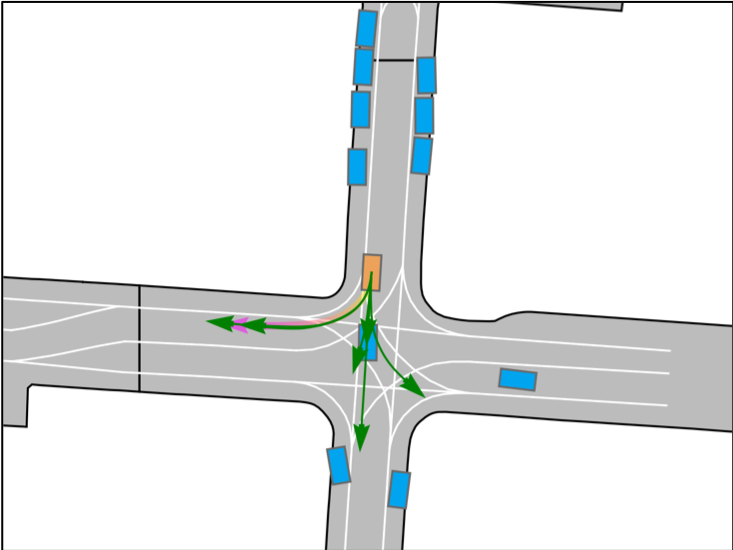}}\vspace{2pt}
                \fbox{
                \includegraphics[width=\linewidth, trim={24pt 22pt 12pt 23pt}, clip]{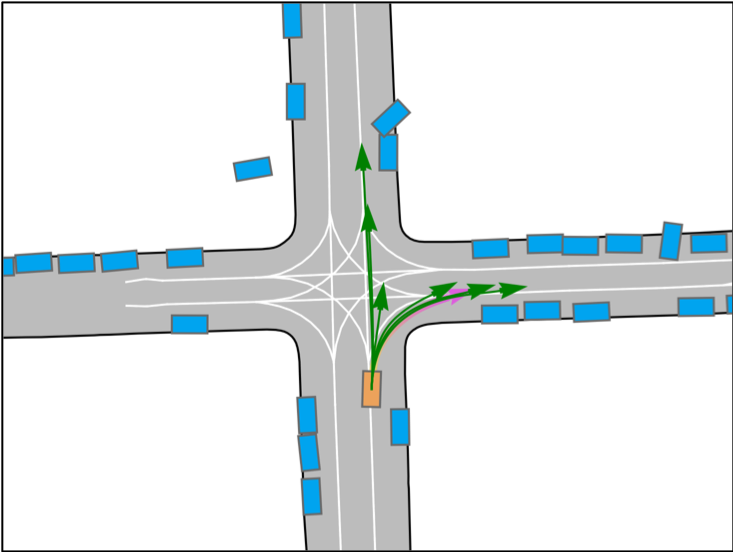}}\vspace{2pt}
                \fbox{
                \includegraphics[width=\linewidth, trim={24pt 22pt 12pt 23pt}, clip]{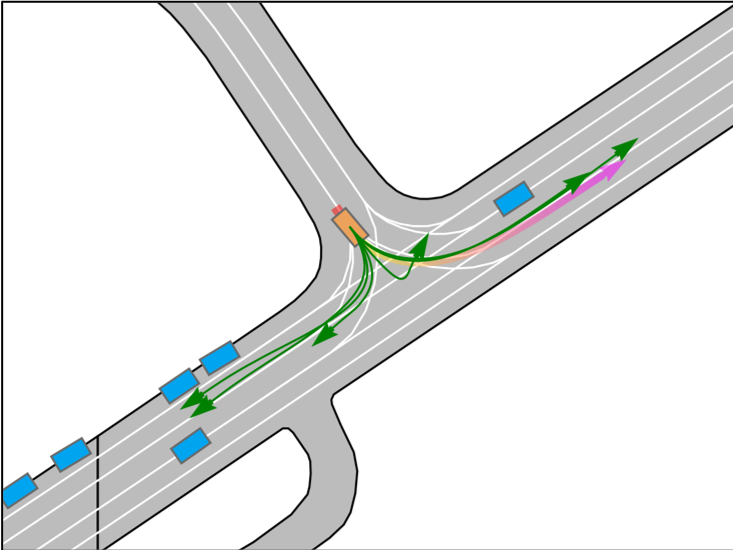}}\vspace{2pt}
                \fbox{
                \includegraphics[width=\linewidth, trim={14pt 22pt 22pt 23pt}, clip]{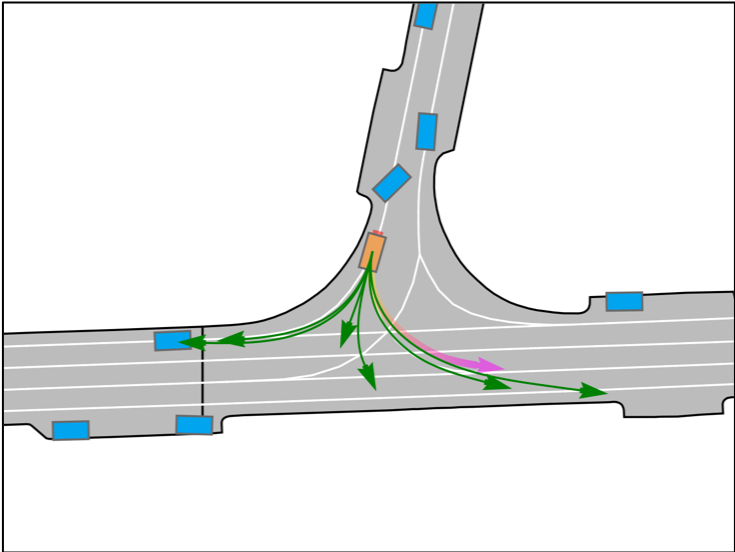}}\vspace{2pt}
                \fbox{
                \includegraphics[width=\linewidth, trim={24pt 22pt 12pt 23pt}, clip]{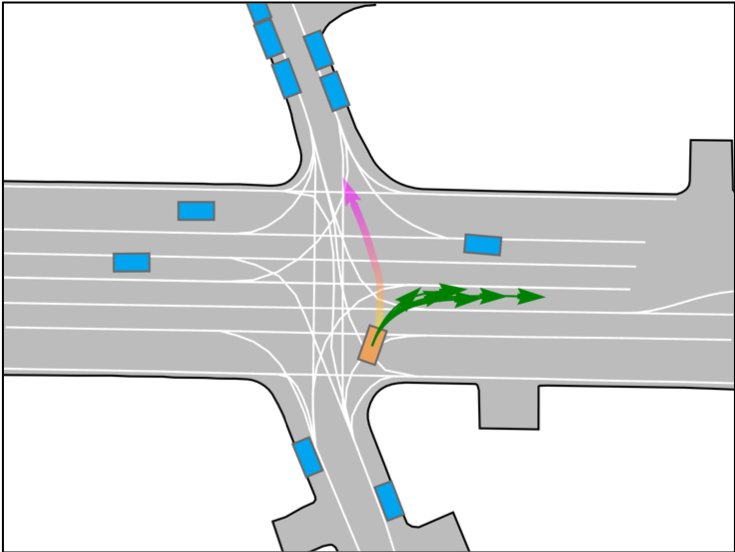}}
			\end{minipage}
		}
		\hfill
		\subfloat[DeMo-CLLS]{
			\begin{minipage}[b]{0.225\textwidth}
                \label{fig.6b}
				\centering
                \fbox{
				\includegraphics[width=\linewidth, trim={13pt 12pt 22pt 23pt}, clip]{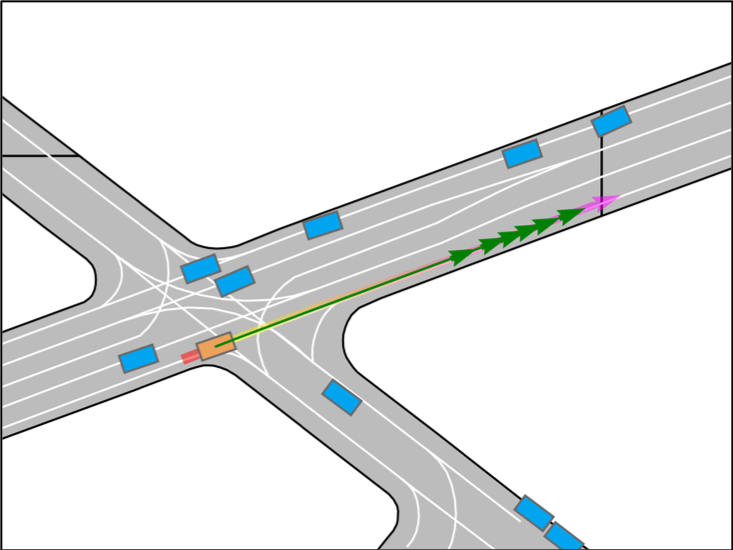}}\vspace{2pt}
                \fbox{
                \includegraphics[width=\linewidth, trim={13pt 12pt 12pt 16pt}, clip]{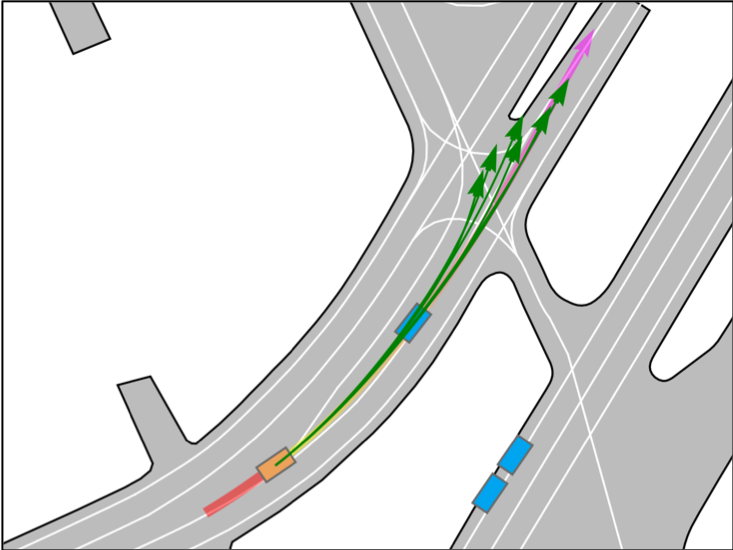}}\vspace{2pt}
                \fbox{
                \includegraphics[width=\linewidth, trim={13pt 12pt 22pt 23pt}, clip]{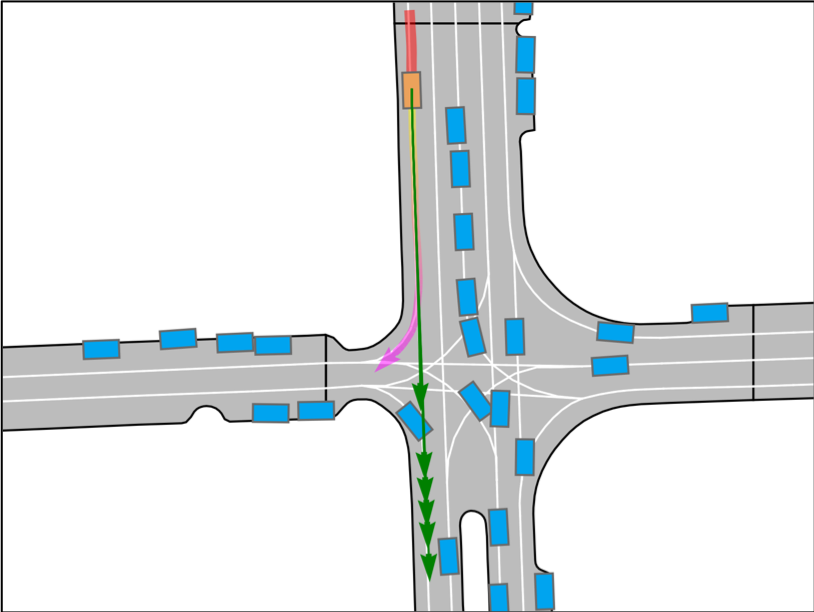}}\vspace{2pt}
                \fbox{
                \includegraphics[width=\linewidth, trim={14pt 12pt 22pt 23pt}, clip]{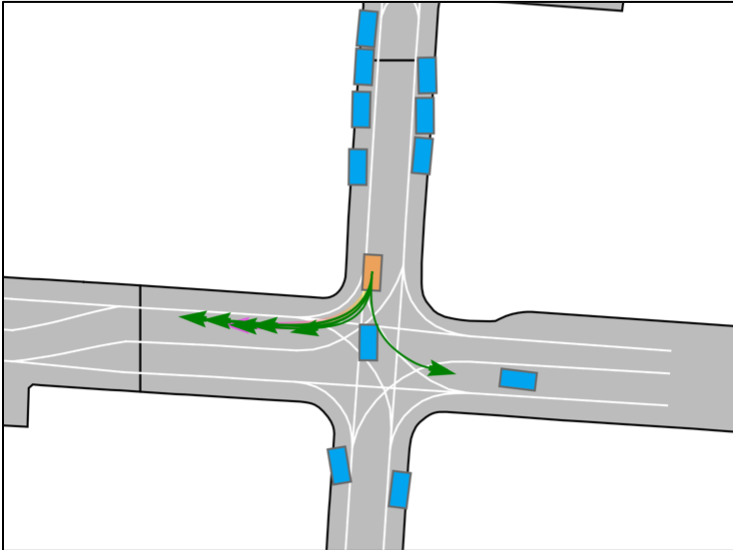}}\vspace{2pt}
                \fbox{
                \includegraphics[width=\linewidth, trim={24pt 22pt 12pt 23pt}, clip]{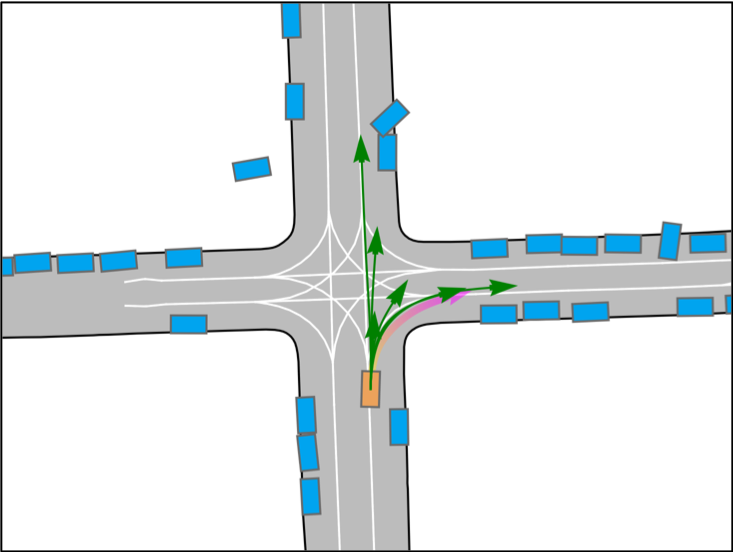}}\vspace{2pt}
                \fbox{
                \includegraphics[width=\linewidth, trim={24pt 22pt 12pt 23pt}, clip]{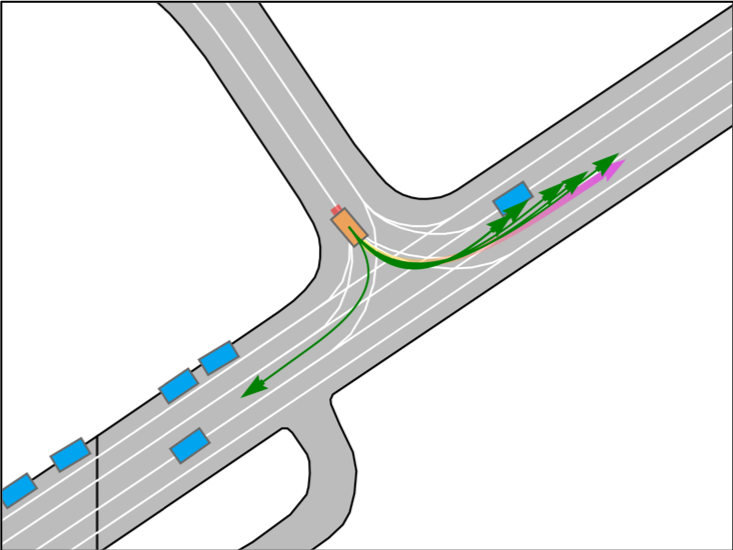}}\vspace{2pt}
                \fbox{
                \includegraphics[width=\linewidth, trim={14pt 22pt 22pt 23pt}, clip]{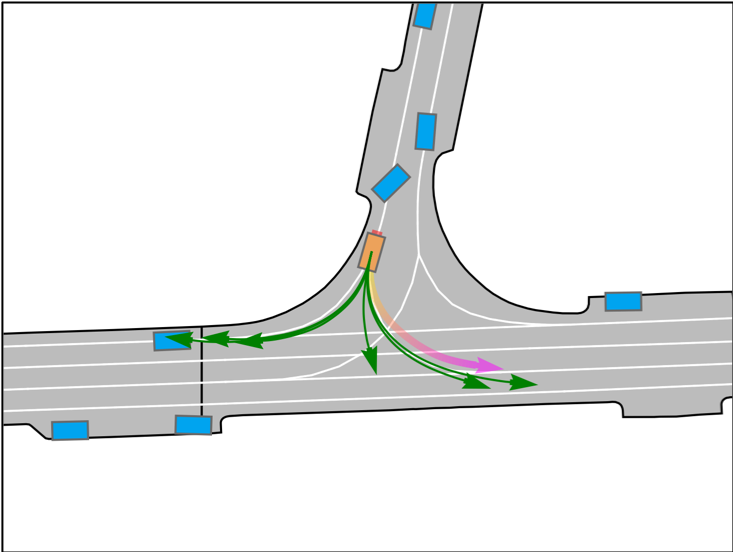}}\vspace{2pt}
                \fbox{
                \includegraphics[width=\linewidth, trim={24pt 22pt 12pt 23pt}, clip]{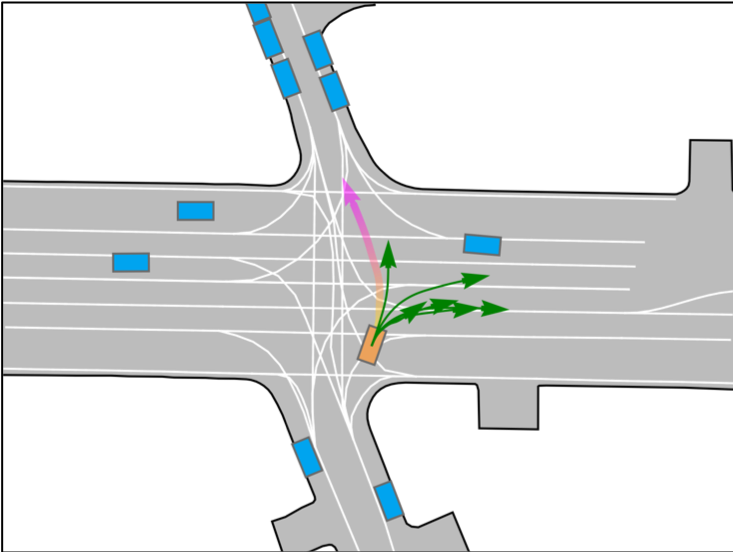}}
			\end{minipage}
		}
		\hfill
		\subfloat[DeMo-PRF (Our)]{
			\begin{minipage}[b]{0.225\textwidth}
                \label{fig.6c}
				\centering
                \fbox{
                \includegraphics[width=\linewidth, trim={13pt 12pt 22pt 23pt}, clip]{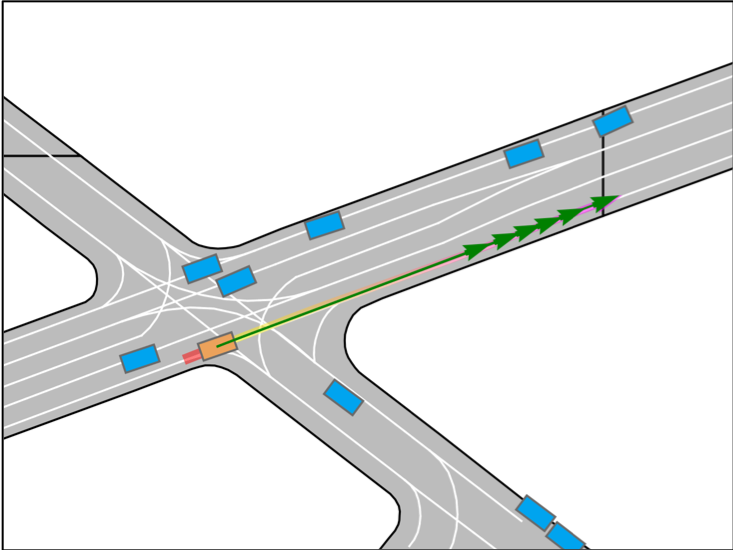}}\vspace{2pt}
                \fbox{
                \includegraphics[width=\linewidth, trim={13pt 12pt 12pt 15pt}, clip]{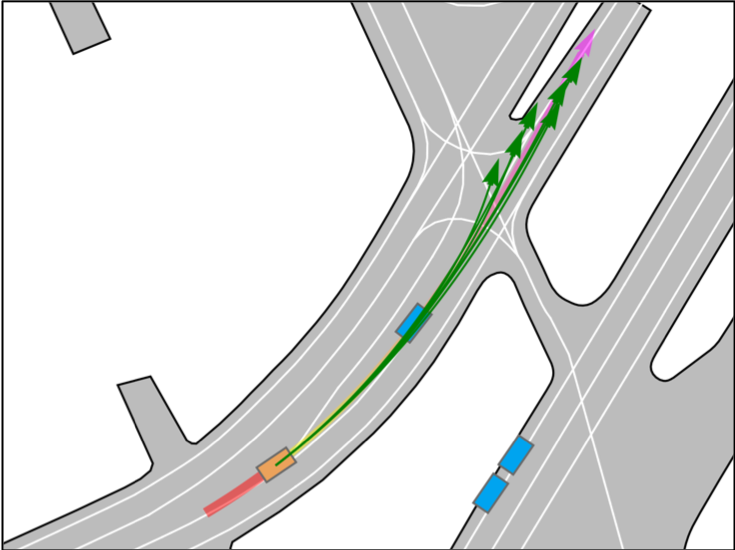}}\vspace{2pt}
                \fbox{
				\includegraphics[width=\linewidth, trim={13pt 12pt 22pt 23pt}, clip]{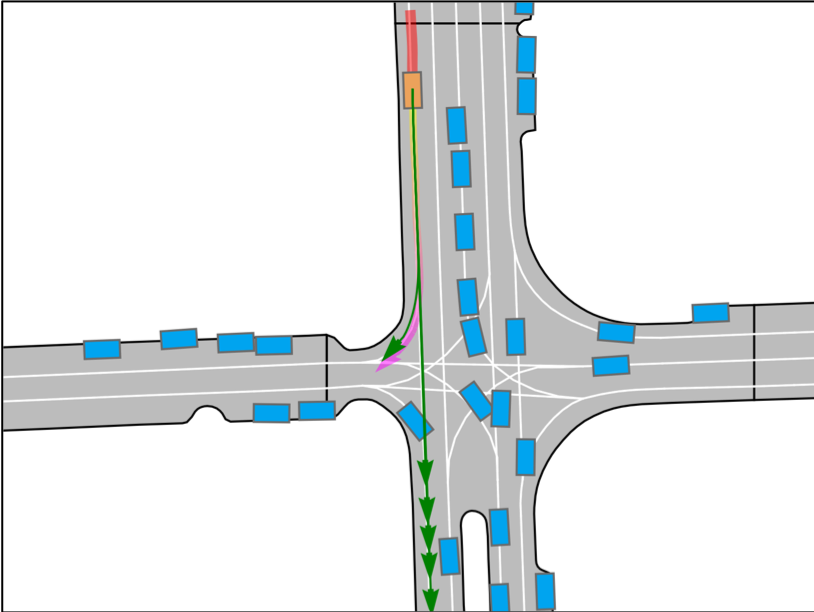}}\vspace{2pt}
                \fbox{
                \includegraphics[width=\linewidth, trim={14pt 12pt 22pt 23pt}, clip]{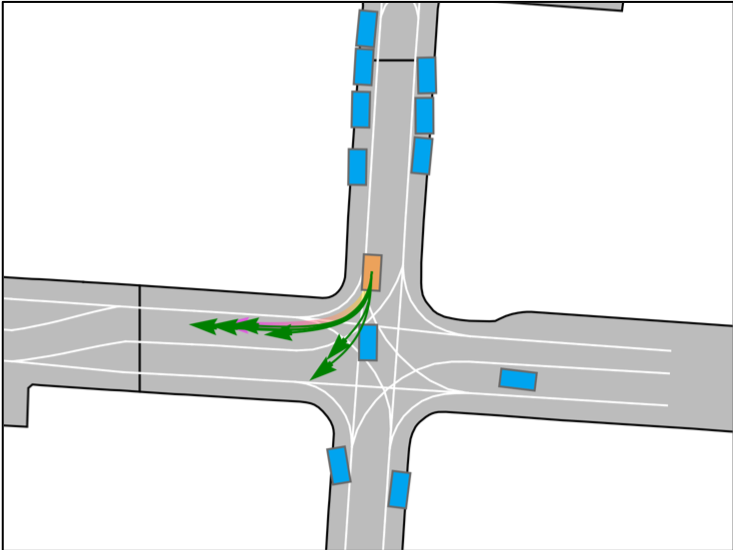}}\vspace{2pt}
                \fbox{
                \includegraphics[width=\linewidth, trim={24pt 22pt 12pt 23pt}, clip]{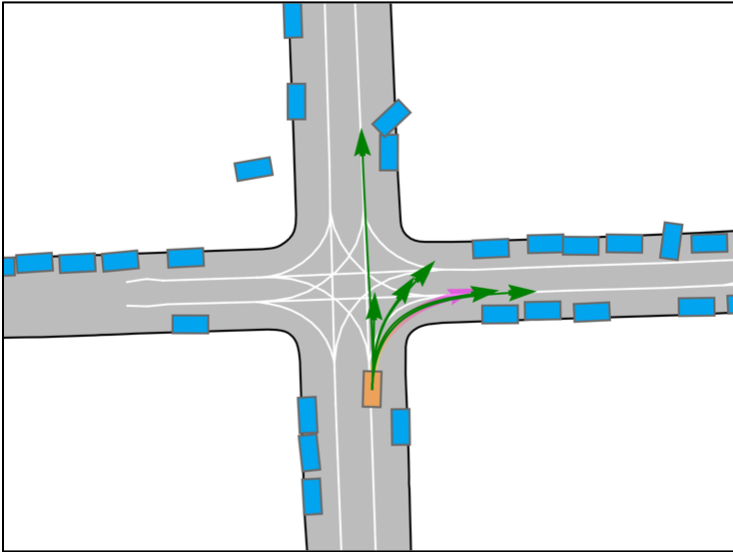}}\vspace{2pt}
                \fbox{
                \includegraphics[width=\linewidth, trim={24pt 22pt 12pt 23pt}, clip]{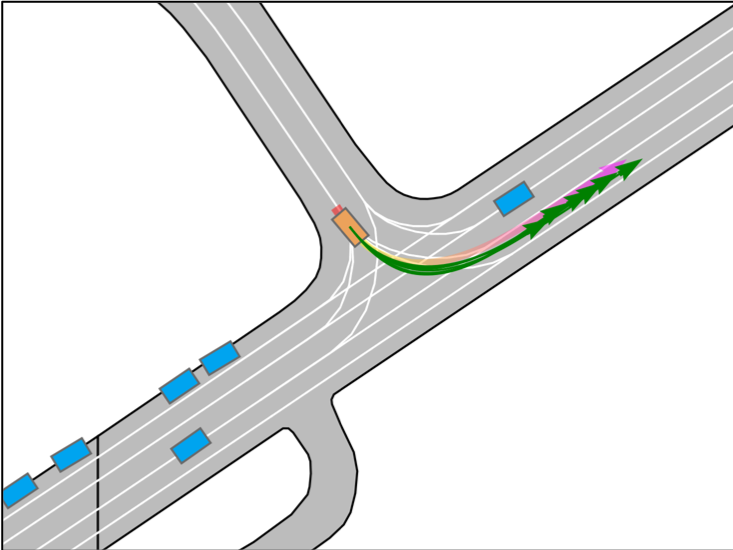}}\vspace{2pt}
                \fbox{
                \includegraphics[width=\linewidth, trim={14pt 22pt 22pt 23pt}, clip]{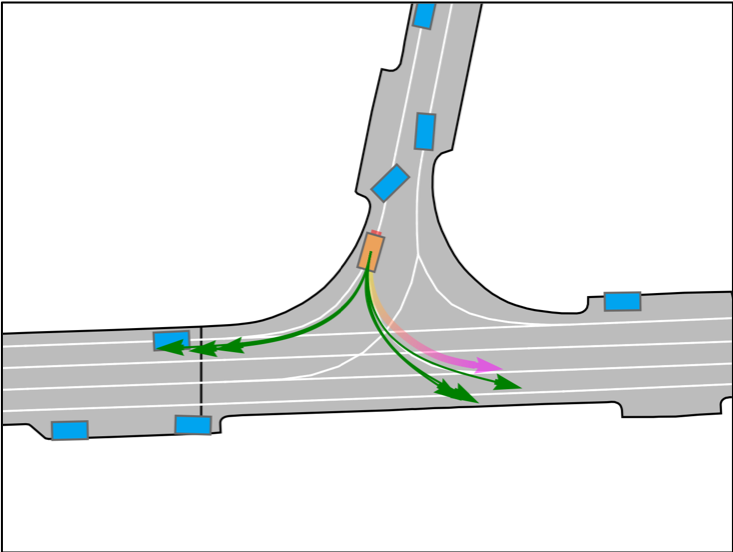}}\vspace{2pt}
                \fbox{
                \includegraphics[width=\linewidth, trim={24pt 22pt 12pt 23pt}, clip]{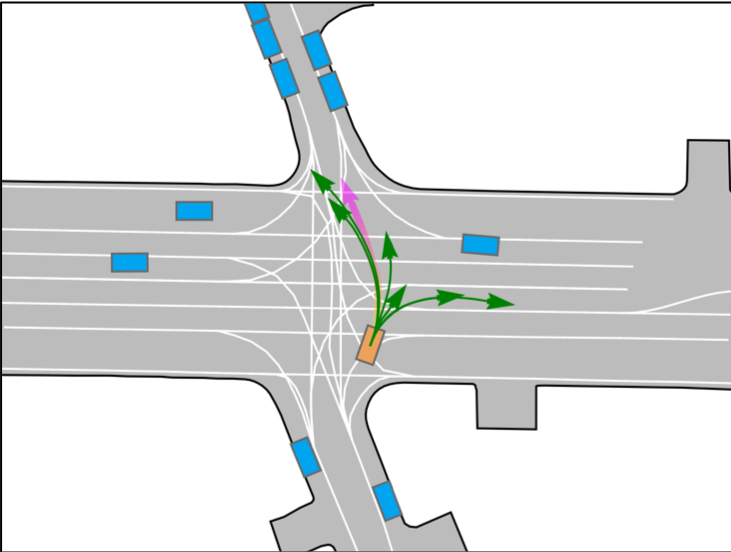}}
			\end{minipage}
		}
		\hfill
		\subfloat[GT]{
			\begin{minipage}[b]{0.225\textwidth}
                \label{fig.6d}
				\centering
                \fbox{
                \includegraphics[width=\linewidth, trim={13pt 12pt 22pt 23pt}, clip]{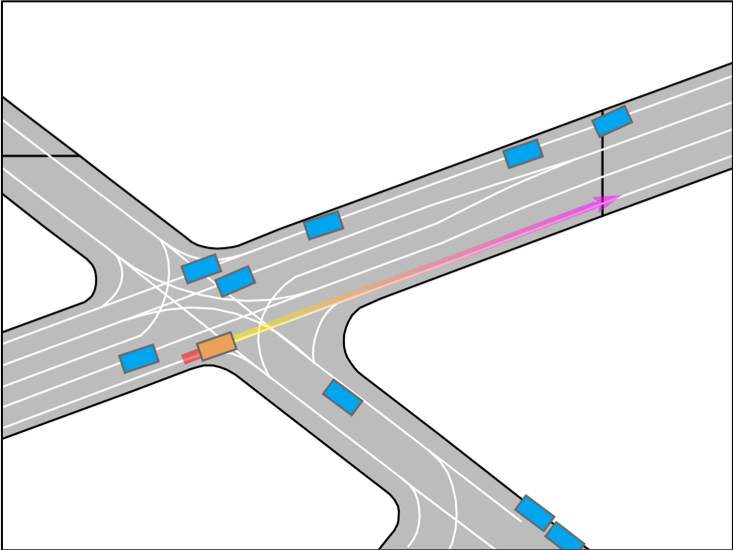}}\vspace{2pt}
                \fbox{
                \includegraphics[width=\linewidth, trim={13pt 12pt 12pt 15pt}, clip]{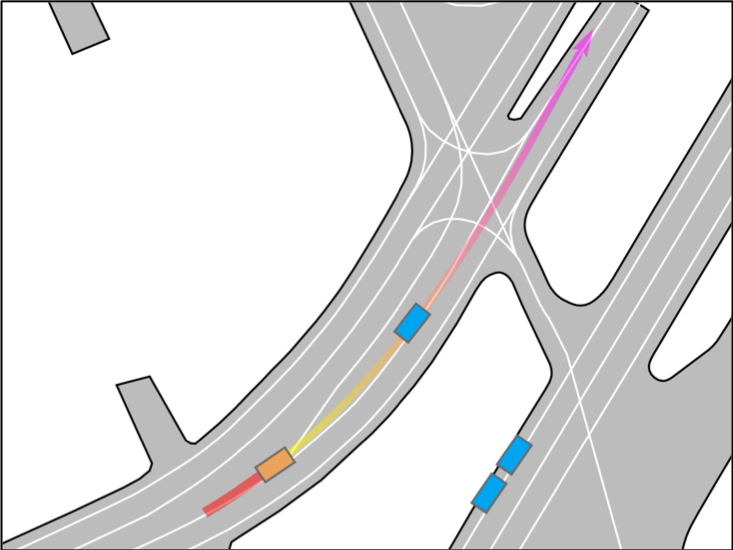}}\vspace{2pt}
                \fbox{
				\includegraphics[width=\linewidth, trim={13pt 12pt 22pt 23pt}, clip]{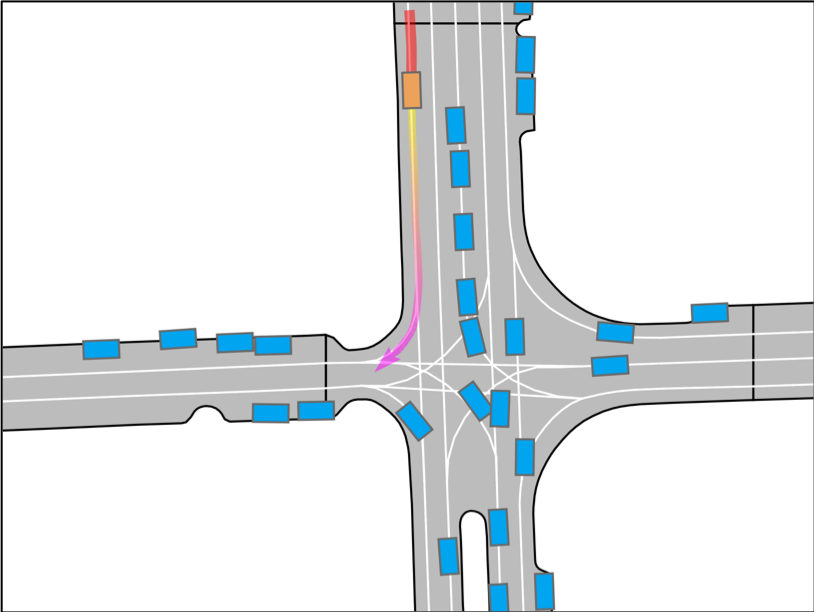}}\vspace{2pt}
                \fbox{
                \includegraphics[width=\linewidth, trim={14pt 12pt 22pt 23pt}, clip]{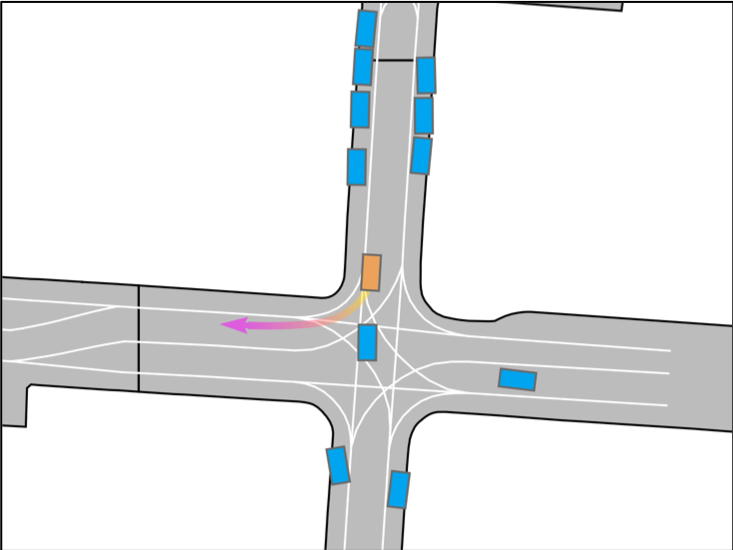}}\vspace{2pt}
                \fbox{
                \includegraphics[width=\linewidth, trim={24pt 22pt 12pt 23pt}, clip]{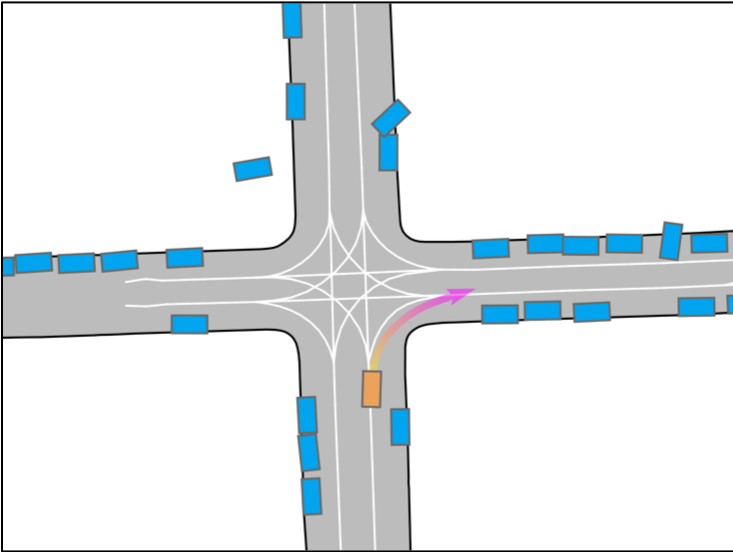}}\vspace{2pt}
                \fbox{
                \includegraphics[width=\linewidth, trim={24pt 22pt 12pt 23pt}, clip]{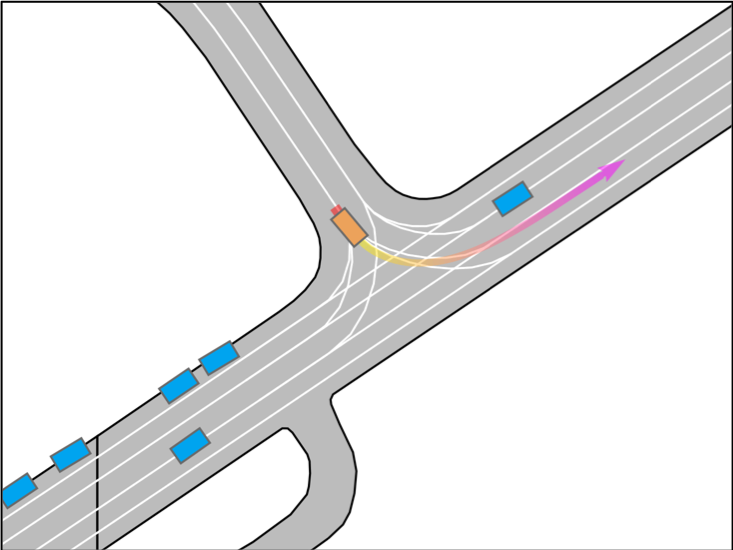}}\vspace{2pt}
                \fbox{
                \includegraphics[width=\linewidth, trim={14pt 22pt 22pt 23pt}, clip]{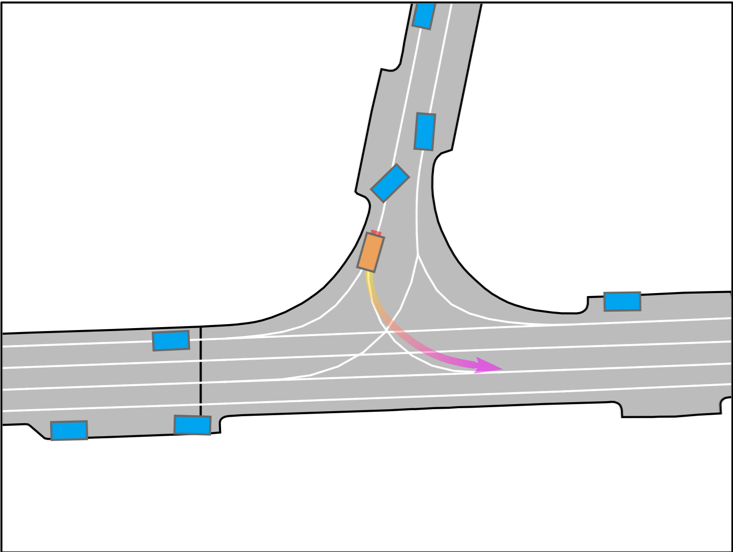}}\vspace{2pt}
                \fbox{
                \includegraphics[width=\linewidth, trim={24pt 22pt 12pt 23pt}, clip]{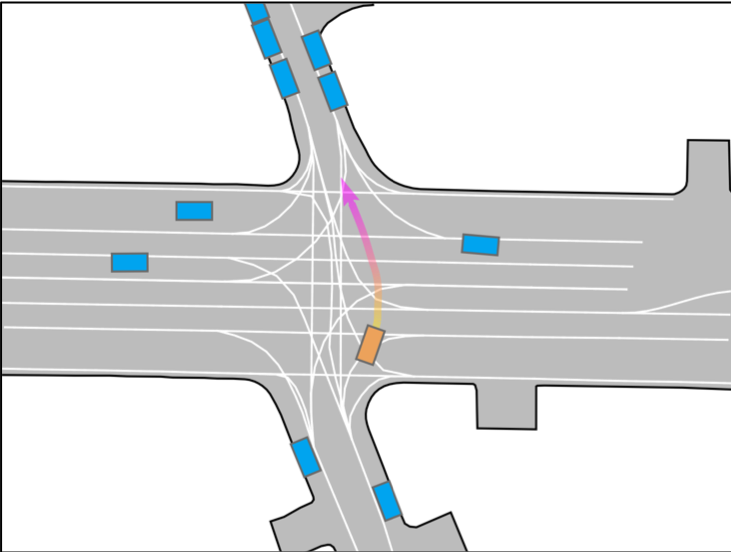}}
			\end{minipage}
		}
	\end{minipage}
    \vspace{-2mm}
	\caption{More qualitative results on the Argoverse~2 validation set. Incomplete observations, predicted trajectories, and ground truth trajectories are shown in yellow, green, and pink, respectively. The absence of an observation trajectory indicates that the vehicle is stationary. Our predictions align more closely with the ground truth compared to other methods.}
    \vspace{-5mm}
	\label{fig:results_vis_ape}
\end{figure*}

\begin{figure*}[!t]
    \centering
    \begin{minipage}[b]{0.85\linewidth}
    \subfloat[Scenario 1]{
        \begin{minipage}[b]{0.225\textwidth}
            \label{fig.7a}
            \centering
            \fbox{
            \includegraphics[width=\linewidth, trim={13pt 12pt 22pt 23pt}, clip]{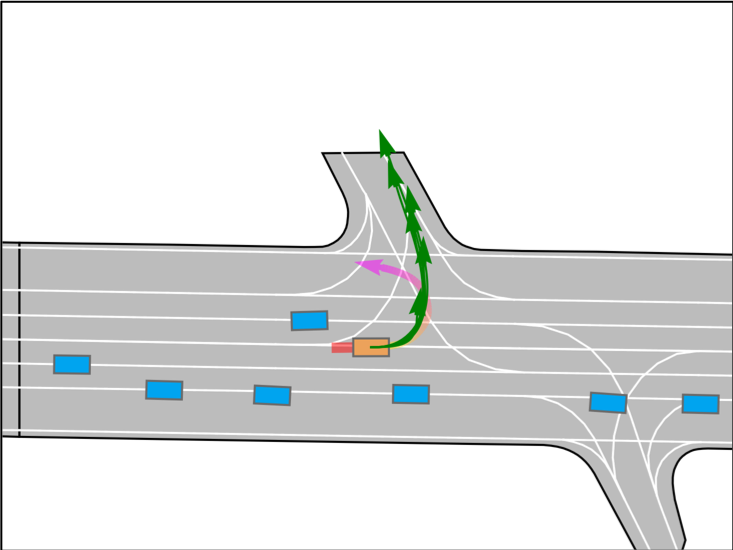}}\vspace{2pt}
            \fbox{
            \includegraphics[width=\linewidth, trim={14pt 12pt 13pt 15pt}, clip]{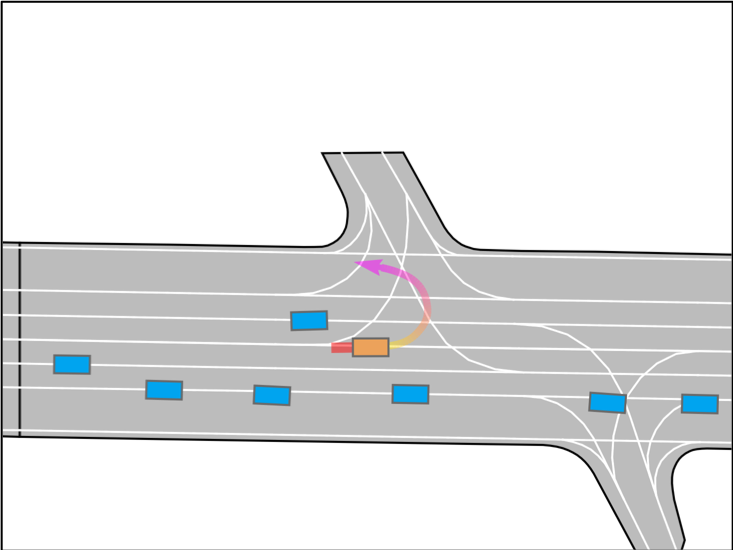}}
        \end{minipage}
    }
    \hfill
    \subfloat[Scenario 2]{
        \begin{minipage}[b]{0.225\textwidth}
            \label{fig.7b}
            \centering
            \fbox{
            \includegraphics[width=\linewidth, trim={13pt 12pt 22pt 23pt}, clip]{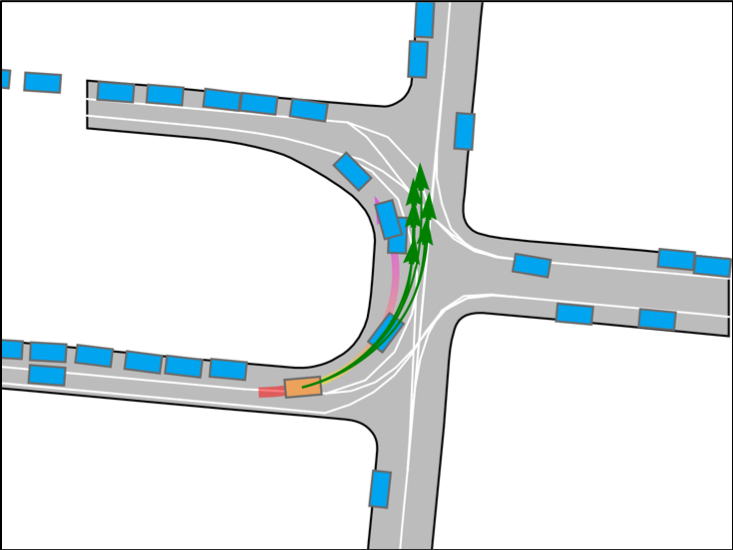}}\vspace{2pt}
            \fbox{
            \includegraphics[width=\linewidth, trim={14pt 12pt 13pt 15pt}, clip]{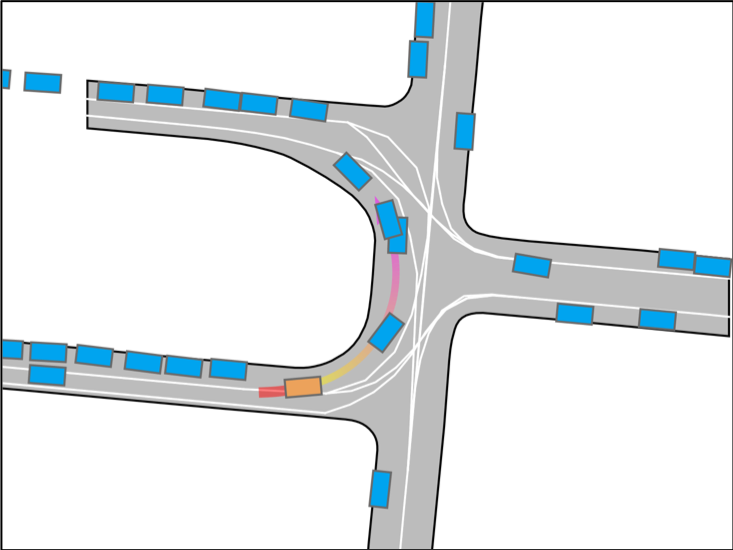}}
        \end{minipage}
    }
    \hfill
    \subfloat[Scenario 3]{
        \begin{minipage}[b]{0.225\textwidth}
            \label{fig.7c}
            \centering
            \fbox{
            \includegraphics[width=\linewidth, trim={13pt 12pt 22pt 23pt}, clip]{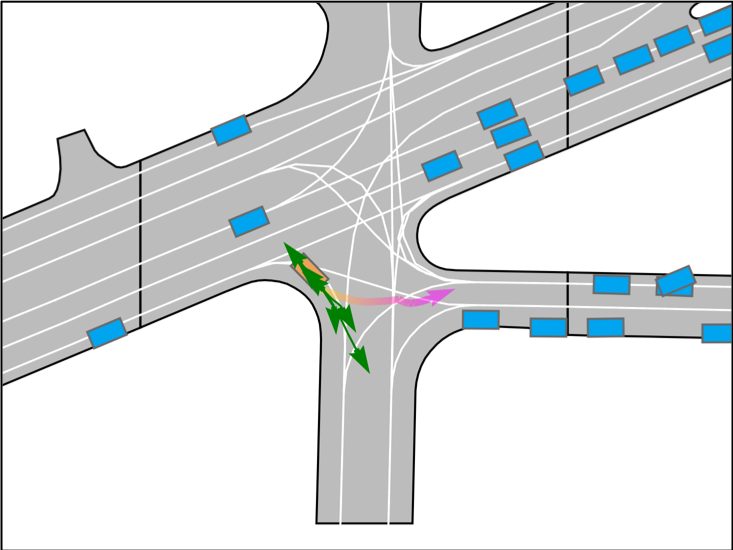}}\vspace{2pt}
            \fbox{
            \includegraphics[width=\linewidth, trim={14pt 12pt 13pt 15pt}, clip]{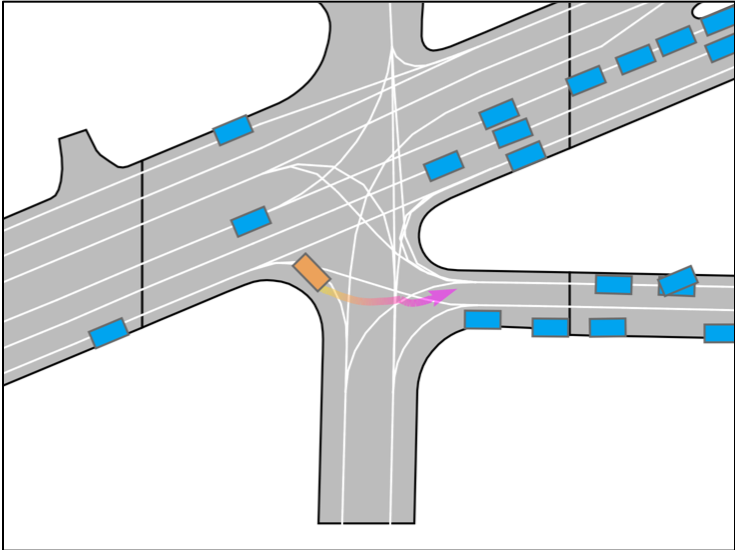}}
        \end{minipage}
    }
    \hfill
    \subfloat[Scenario 4]{
        \begin{minipage}[b]{0.225\textwidth}
            \label{fig.7d}
            \centering
            \fbox{
            \includegraphics[width=\linewidth, trim={14pt 12pt 13pt 16pt}, clip]{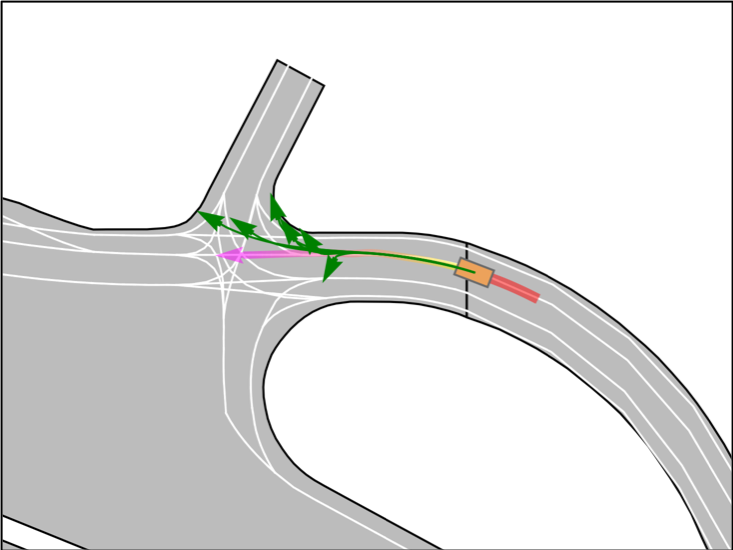}}\vspace{2pt}
            \fbox{
            \includegraphics[width=\linewidth, trim={14pt 12pt 13pt 15pt}, clip]{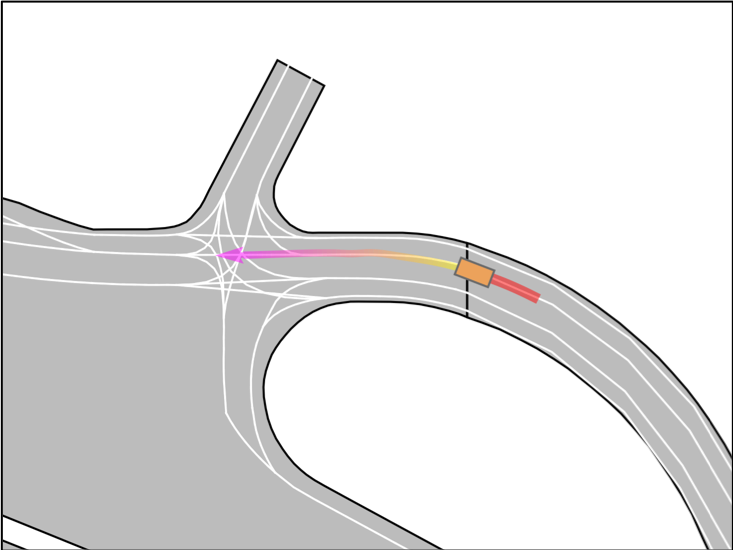}}
        \end{minipage}
    }
    \end{minipage}
    \vspace{-2mm}
	\caption{Failure cases of DeMo-PRF on the Argoverse~2 validation set. The first and second rows visualize the predicted trajectories and ground-truth trajectories, respectively. Incomplete observations, predicted trajectories, and ground truth trajectories are shown in yellow, green, and pink, respectively. The absence of an observation trajectory indicates that the vehicle is stationary.}
    \vspace{-5mm}
	\label{fig:failure_cases}
\end{figure*}

\vspace{-1mm}
\section{Appendix for Loss Functions}
\vspace{-1mm}
\label{sec:losses_ape}

A smooth-L1 loss and a cross-entropy loss are employed to train the decoder and RPM, as introduced in Section~\ref{sec:}. The ground-truth future trajectories, predicted future trajectories, and their probability are represented by $\mathbf{Y} \in \mathbb{R}^{N_a \times T_f \times 2}$, $\tilde{\mathbf{Y}} \in \mathbb{R}^{N_a \times K \times T_f \times 2}$, and $\mathbf{P} \in \mathbb{R}^{N_a \times K}$, where $N_a$, $K$, $T_f$, and $2$ represents the number of predicted agents, the number of predicted modes, the prediction horizon, and the coordinate dimensions, respectively. These variables are used to compute the smooth-L1 loss and cross-entropy loss.

\noindent\textbf{Smooth-L1 regression loss.} The smooth-L1 regression loss is computed using the ground-truth future trajectories $\mathbf{Y}$ and predicted future trajectories $\tilde{\mathbf{Y}}$ as follows:
\[
\mathcal{L}_{\text{reg}}
= \frac{1}{N_a T_f}
  \sum_{i=1}^{N_a} \sum_{t=1}^{T_f} \operatorname{SmoothL1}\bigl(\tilde{\mathbf{Y}}_{i,k_i^\star,t} - \mathbf{Y}_{i,t}\bigr)
\]
where $k_i^\star$ denotes the index of the best predicted mode for agent $i$.

\noindent\textbf{Cross-entropy classification loss.}
For probability score classification, the index $k_i^\star$, corresponding to the mode with the smallest ADE of agent $i$, is used as the ground-truth class label. Then, using the predicted probability $\mathbf{P}$ and the ground-truth class label, the classification loss is calculated as:
\begin{equation}
    \mathcal{L}_{\text{cls}} = - \frac{1}{N_a} \sum_{i=1}^{N_a} \log \mathbf{P}_{i,k_i^\star}.
\end{equation}

The overall training loss for the decoder and RPM is the sum of the regression and classification terms.

\vspace{-1mm}
\section{Appendix for Evaluation Metrics}
\vspace{-1mm}
\label{sec:metrics_ape}

We adopt commonly used metrics, namely mADE$_K$, mFDE$_K$, b-mFDE$_K$, and MR$_K$, to evaluate PRF, as described in Section~\ref{sec:exp_settings}. The ground-truth future trajectories $\mathbf{Y}$, predicted future trajectories $\tilde{\mathbf{Y}}$, and their associated probabilities $\mathbf{P}$ are used to compute these metrics. Specifically, for each agent $i$ and mode $k$, the ADE and FDE are defined as follows:
\begin{equation}
\begin{aligned}
    \mathrm{ADE}_{i,k} &= \frac{1}{T_f} \sum_{t=1}^{T_f} \left\lVert \tilde{\mathbf{Y}}_{i,k,t} - \mathbf{Y}_{i,t} \right\rVert_2, \\
    \mathrm{FDE}_{i,k} &= \left\lVert \tilde{\mathbf{Y}}_{i,k,T_f} - \mathbf{Y}_{i,T_f}\right\rVert_2.
    \end{aligned}
\end{equation}
Then, mADE$_K$ and mFDE$_K$ are calculated as the average minimum ADE and FDE over $K$ modes, respectively:
\begin{equation}
    \begin{aligned}
        \mathrm{mADE}_K &= \frac{1}{N_a} \sum_{i=1}^{N_a} \min_{1 \le k \le K} \mathrm{ADE}_{i,k},\\
        \mathrm{mFDE}_K &= \frac{1}{N_a} \sum_{i=1}^{N_a} \min_{1 \le k \le K} \mathrm{FDE}_{i,k}.
    \end{aligned}
\end{equation}
The $\mathrm{b\text{-}mFDE}_K$ metric augments $\mathrm{mFDE}_K$ with a Brier-style penalty based on the probability of the best predicted mode:
\begin{equation}
    \mathrm{b\text{-}mFDE}_K = \frac{1}{N_a} \sum_{i=1}^{N_a}
    \left[
        \mathrm{FDE}_{i,k_i^\star}
        + \bigl( 1 - P_{i,k_i^\star} \bigr)^2
    \right].
\end{equation}
The $\mathrm{MR}_K$ metric measures the fraction of agents for which even the best of the $K$ predicted trajectories deviates from the ground truth by more than a threshold $\delta = 2.0$ meters at the final time step:
\begin{equation}
    \mathrm{MR}_K = \frac{1}{N_a} \sum_{i=1}^{N_a}
    \mathbf{1} \Bigl(
        \mathrm{FDE}_{i,k_i^\star} > \delta
    \Bigr),
\end{equation}
where $\mathbf{1}(\cdot)$ is the indicator function that returns $1$ if the condition is true and $0$ otherwise. These metrics can be extended to the entire dataset by averaging over the total number of predicted agents across all scenes.

\vspace{-1mm}
\section{Appendix for Qualitative Evaluations}
\vspace{-1mm}
\label{sec:qualit_ape}

Additional qualitative results, complementing those presented in Fig.~\ref{fig:results_vis}, are shown in Fig.~\ref{fig:results_vis_ape}. All results are predicted from very short observation horizons of only 10 timesteps. In some scenarios, the absence of an observation trajectory indicates that the vehicle is stationary during the observation window. These qualitative results, spanning various driving scenarios, further highlight the state-of-the-art performance of the proposed PRF.

\noindent \textbf{Failure case.} Fig.~\ref{fig:failure_cases} illustrates four failure cases when using very short observation horizons of only 10 timesteps. Fig.~\ref{fig.7a} and Fig.~\ref{fig.7b} depict failure cases in U-turn scenarios. Fig.~\ref{fig.7c} shows a failure case in a compound turn scenario, while Fig.~\ref{fig.7d} presents a failure case in the on-ramp merging scenario. These scenarios present long-tail problems for trajectory prediction, even with complete observation lengths. With such short and incomplete observations, the proposed PRF initially tracks the ground-truth motion but eventually deviates as the maneuver becomes more complex. To improve predictions in these scenarios, future work could focus on enhancing the modeling of interactions among multiple agents and incorporating additional high-level context, such as traffic signals and right-of-way rules, as structural constraints on the predicted trajectories.

\begin{figure}[!t]
    \centering
    \includegraphics[width=0.95\linewidth,trim={10pt, 10pt, 10pt, 10pt}, clip]{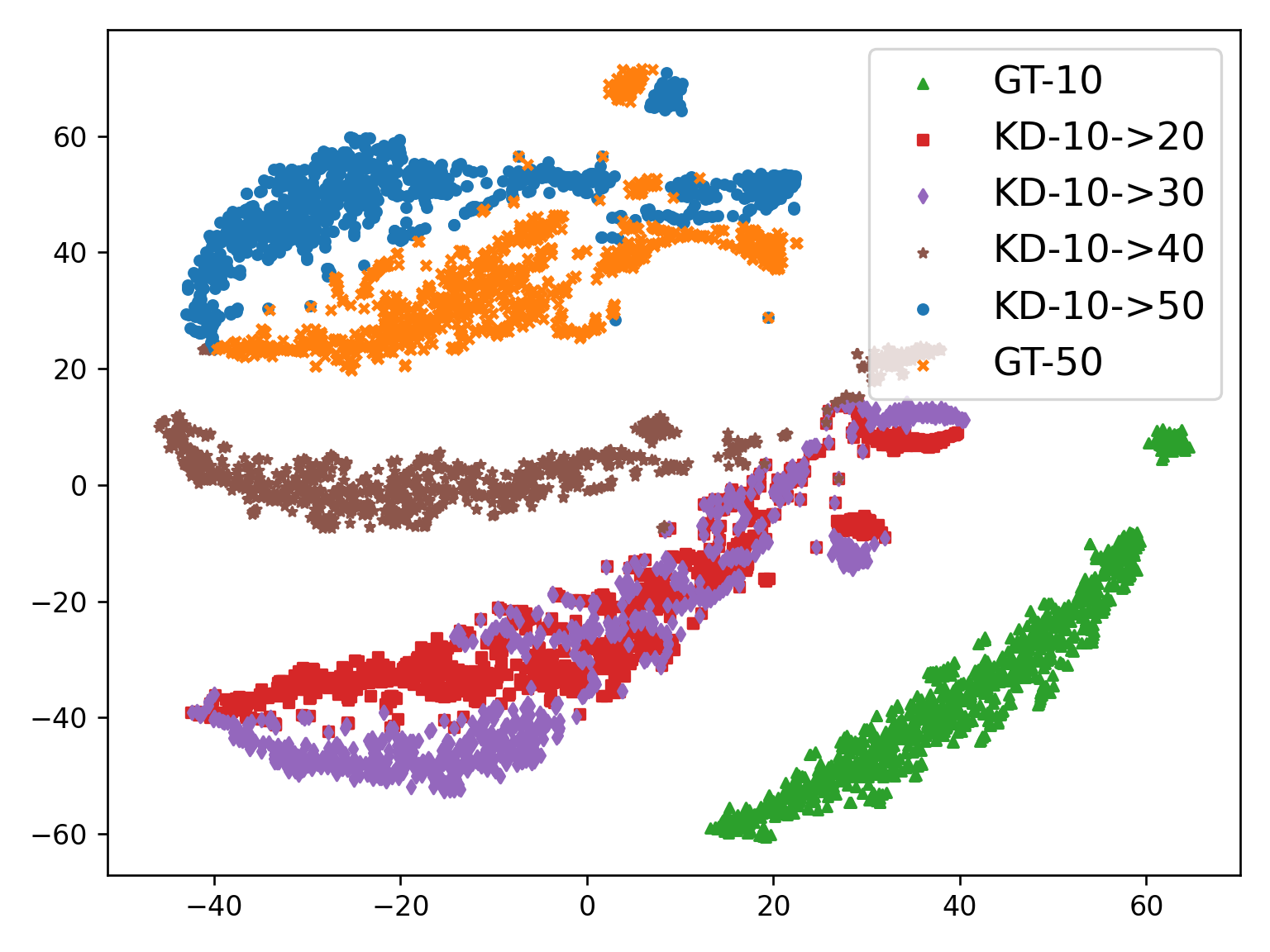}
    \vspace{-2mm}
    \caption{t-SNE visualization of features distilled by the progressive strategy. Green and orange points represent features extracted from trajectories with observation lengths of 10 and 50 timesteps, respectively. Red, purple, brown, and blue points correspond to features distilled from 10-step observations to those with 20, 30, 40, and 50 steps, which gradually shift from the manifold of 10-step observations toward that of 50-step observations.}
    \label{fig:tsne-ape}
    \vspace{-5mm}
\end{figure}

\vspace{-1mm}
\section{Appendix for Interpretability Analysis}
\vspace{-1mm}
\label{sec:inter_ape}

Additional interpretability analysis, complementary to Fig.~\ref{fig:tsne-comparison}, is presented in Fig.~\ref{fig:tsne-ape}. This figure visualizes the t-SNE of features extracted from observation lengths of 10 and 50 timesteps, as well as features distilled from 10-step observations to those with 20, 30, 40, and 50 timesteps. 
The visualization shows that, as progressive distillation proceeds, features distilled from trajectories with an observation length of 10 timesteps gradually converge toward the features obtained from 50-step observations. This demonstrates that decomposing direct distillation into a sequence of progressive distillation steps reduces the difficulty of the distillation process and effectively distills representations from short trajectories into those of complete trajectories.


\end{document}